\documentclass[12pt]{ociamthesis}
\pdfoutput=1

\usepackage{etex}
\usepackage{times}

\usepackage{booktabs}
\usepackage{mathtools}
\usepackage{amsfonts}
\usepackage{amssymb}
\usepackage{stmaryrd}
\SetSymbolFont{stmry}{bold}{U}{stmry}{m}{n}

\usepackage{multirow}
\usepackage{subcaption}
\usepackage{array}
\usepackage[outercaption]{sidecap}
\usepackage{dcolumn}

\usepackage{url}
\usepackage{graphicx}
\usepackage[export]{adjustbox}
\usepackage{xspace}
\usepackage[table]{xcolor}

\usepackage{tikz}
\usepackage{pgfplots}
\usepackage{tikz-qtree}
\usetikzlibrary{arrows,positioning,calc,backgrounds,shapes,fit,mindmap,shadows}
\usetikzlibrary{matrix,decorations.pathreplacing}
\usetikzlibrary{fadings}

\usepackage{thesis}

\makeatletter
\newcommand{\@BIBLABEL}{\@emptybiblabel}
\newcommand{\@emptybiblabel}[1]{}
\makeatother
\usepackage{hyperref}

\hypersetup{
    colorlinks=true,
    linkcolor=black,
    citecolor=black,
    filecolor=black,
    urlcolor=black,
}

\DeclareMathOperator*{\argmax}{arg\,max}
\DeclareMathOperator*{\argmin}{arg\,min}

\title{Scalable Machine Translation in\\[1ex] Memory Constrained Environments}

\author{Paul-Dan Baltescu}
\college{St. Hugh's College}

\degree{Master by Research}
\degreedate{Trinity 2016}

\begin{document}

\setcounter{secnumdepth}{3}
\setcounter{tocdepth}{2}

\maketitle

\begin{abstract}
Machine translation is the discipline concerned with developing automated tools for translating from one human language to another. Statistical machine translation (SMT) is the dominant paradigm in this field. In SMT, translations are generated by means of statistical models whose parameters are learned from bilingual data. Scalability is a key concern in SMT, as one would like to make use of as much data as possible to train better translation systems.

In recent years, mobile devices with adequate computing power have become widely available. Despite being very successful, mobile applications relying on NLP systems continue to follow a client-server architecture, which is of limited use because access to internet is often limited and expensive. The goal of this dissertation is to show how to construct a scalable machine translation system that can operate with the limited resources available on a mobile device. 

The main challenge for porting translation systems on mobile devices is me\-mory usage. The amount of memory available on a mobile device is far less than what is typically available on the server side of a client-server application. In this thesis, we investigate alternatives for the two components which prevent standard translation systems from working on mobile devices due to high memory usage. We show that once these standard components are replaced with our proposed alternatives, we obtain a scalable translation system that can work on a device with limited memory.

The first two chapters of this thesis are introductory. \autoref{chapter:intro} discusses the task we undertake in greater detail and highlights our contributions. \autoref{chapter:smt} provides a brief introduction to statistical machine translation.

In \autoref{chapter:extractors}, we explore online grammar extractors as a memory efficient alternative to phrase tables. We propose a faster and simpler extraction algorithm for translation rules containing gaps, thereby improving the extraction time for hierarchical phase-based translation systems.

In \autoref{chapter:nlms}, we conduct a thorough investigation on how neural language models should be integrated in translation systems. We settle on a novel combination of noise contrastive estimation and factoring the output layer using Brown clusters. We obtain a high quality translation system that is fast both when training and decoding and we use it to show that neural language models outperform traditional n-gram models in memory constrained environments.

\autoref{chapter:conclusions} concludes our work showing that online grammar extractors and neural language models allow us to build scalable, high quality systems that can translate text with the limited resources available on a mobile device.
\end{abstract}

\begin{romanpages}
\tableofcontents
\listoffigures
\listoftables
\end{romanpages}

\baselineskip=20pt plus1pt
\chapter{Introduction}\label{chapter:intro}
Machine translation is the discipline concerned with developing automated processes for translating from one natural (human) language to another. Although machine translation has been a subject of interest in the research community since the late 1940s \cite{Weaver1949}, the field has only seen significant progress since the 1990s with the rise in popularity of statistical methods for translation \cite{Brown1993}. Statistical machine translation makes use of parallel data to automatically infer rules which can later be combined to produce translations of input sentences. As with most natural language systems employing statistical models, the quality of translation systems increases as more data is used to train the underlying models. Scalability is a key concern in machine translation, as one would like to use as much data as possible when building a translation system.

In the last decade, mobile devices with adequate computing power (e.g., smartphones, tablets, etc.) have become widely available and have started playing an important role in the daily life of millions of people. Mobile applications using NLP systems such as speech recognizers or translation systems have been incredibly successful because they lower the barrier for accessing information on the go. Most of these applications have a client-server architecture, where the heavy computation specific to NLP tasks is done on the server side, while the client side is only used to render the user interface of the application. The main bottleneck of this approach is access to internet, which is often limited and expensive. To work around this problem, one has to construct NLP systems that can operate with the limited resources available on mobile devices. The key limitation is the amount of memory available, typically limited to 1GB, which is 2-3 orders of magnitude less than what is usually available on the server side of a client-server application. A similar problem is encountered when developing NLP systems that are expected to run on commodity machines. Despite having slightly more memory than mobile devices, average home computers are still far less powerful than the high-end machines used by software companies or research institutes. In this thesis, we seek to develop translation systems that work on memory constrained devices.

\section{Thesis Goals}

The primary aim of this thesis is to present a scalable approach for constructing high quality machine translation systems that can run in memory constrained environments such as mobile devices or commodity machines. We tackle the two main challenges that prevent standard translation systems from working in such environments: the representation of the translation model in memory and structure of the language model. First, we explore compact representations of translation models which rely on suffix arrays to efficiently locate phrases in the source side of a parallel corpus and extract translation rules on the fly. Second, we investigate neural language models as a memory efficient alternative to traditional n-gram language models and analyze the effect of the most popular scaling techniques for neural models on end-to-end translation quality. We show that by introducing our proposed alternatives in a standard translation system to replace their equivalent components, we obtain a fast, compact and high quality translation system.

Throughout this thesis, we seek to evaluate the approaches we propose using several metrics. Above all, we are interested in producing high quality translation systems and we follow the standard practice of reporting BLEU scores \cite{Papineni2002}. We also report the amount of memory needed to store the models we investigate, as it is our goal to show that these models are compact enough to be used in memory constrained environments. Finally, we must ensure that our models are fast enough to be practical, both when training a translation system and when using it to translate new sentences. We achieve this by keeping track of the time needed to train each component individually and of the average time needed to decode a test sentence. We note that building a translation system is a time consuming task (which may take up to several days), but we are not interested to perform this task with the limited resources available on the client device. Instead, we would like to train our models on a powerful machine and download them on the client device when access to internet is available, leaving decoding as the only operation to be performed on the client. 

In our dissertation, we also explore decisions which lead to trade-offs between these metrics. For example, certain scaling techniques for neural models result in higher translation quality, but make decoding slower, or depending on the amount of memory available, using either neural language models or back-off n-gram models will lead to higher BLEU scores.

\section{Contributions}

In this section, we summarize the main contributions of this thesis.

In Chapter \ref{chapter:extractors}, we discuss compact alternatives to phrase tables, the traditional data structures used to represent translation models in memory. We highlight the importance of these alternatives, in particular in the context of hierarchical phrase-based translation systems \cite{Chiang2007}. We choose online grammar extractors as the basis of our work and provide supporting arguments to motivate the decision. We present an efficient implementation of an online grammar extractor using suffix arrays based on \newcite{Lopez2007}. We introduce a novel algorithm for extracting hierarchical translation rules with significantly lower running time. Our approach is also much simpler to implement than \newcite{Lopez2007}.

In Chapter \ref{chapter:nlms}, we conduct a thorough analysis on integrating neural language models in translation systems. Although the idea of incorporating these models in MT is not new, we are the first to explore it with the goal of producing a compact translation system and to focus on the properties of neural language models as the sole language models in the system. The latter is important because our hypothesis is that most of the language modeling is otherwise done by the back-off n-gram model, with the neural language model only acting as a differentiating factor when the n-gram model cannot provide a decisive probability. We show that neural language models clearly outperform traditional n-gram models in memory constrained environments, but when the memory restriction is lifted, back-off n-gram models are more effective than their neural counterparts. Scaling neural language models is a difficult task, but crucial for obtaining practical translation systems. We investigate the impact of several frequently used scaling techniques on end-to-end translation quality. We discover that a novel combination of noise contrastive estimation \cite{Mnih2012} and factoring the softmax layer using Brown clusters \cite{Brown1992} is the most pragmatic solution for efficient training and decoding with neural language models. Finally, we explore two extensions to neural language models (one investigated for the first time in the context of translation systems) with the goal of boosting translation quality further.

In \autoref{chapter:conclusions}, we show that by combining the techniques introduced in the earlier chapters, we obtain a high quality system that fits within the 1 GB memory constraint. We evaluate our system on three language pairs and show that it outperforms a traditional system trained on sampled data to match the memory requirements by 0.7-1.5 BLEU points. The proposed techniques are scalable both when training the model and when using it to decode new sentences.

Another important contribution of our thesis is that we open source our code and make it easy to integrate with the most popular translation toolkits. Our suffix array grammar extractor\footnote{The code has been released at: \url{https://github.com/redpony/cdec/tree/master/extractor}.} is released as part of cdec \cite{Dyer2010} and has been included as part of the default instructions for building a baseline system with the toolkit. The extractor is designed as a standalone tool, and in order to incorporate it in other translation toolkits, one has to write only the new interface between the translation system and the grammar extractor. We also release OxLM, a scalable neural language modeling toolkit\footnote{The language modeling toolkit is available at: \url{https://github.com/pauldb89/oxlm}.}. The models trained with OxLM can be integrated as features in the cdec \cite{Dyer2010} and Moses \cite{Koehn2007} decoders. In contrast to other open source neural language toolkits for MT \cite{Vaswani2013}, we allow our models to be explicitly normalized. This is crucial for obtaining high quality translation systems when additional back-off n-gram models cannot be used due to memory constraints. Also, unlike \newcite{Schwenk2010}, our models can be integrated directly in the decoder. Also, we do not use a backup n-gram model to score rare words, because it is not feasible to do so with limited memory.
    
\section{Thesis Structure}

In this section, we discuss the structure of the thesis and summarize the contents of each chapter. Part of the work presented in this dissertation is based on publications of which we are the main author. We indicate which parts rely on previously published material, as we overview the topics covered by each chapter.

\begin{description}
  \item[Chapter \ref{chapter:smt}: Statistical Machine Translation] \hfill \vspace{0.25cm} \\    
	This chapter presents a brief introduction to statistical machine translation. Our goal is to explain how a translation system works, as we prepare the reader for the topics covered in the next chapters. We discuss the standard approaches employed by each component of a translation system and show where difficulties arise when the amount of memory is limited. We compare two formalisms that lie at the foundation of most translation systems in use today: finite state transducers and synchronous context free grammars. We focus our exposition on the topics relevant for reaching the goal of constructing a compact translation system and refer the interested reader to \newcite{Lopez2008b} for a thorough review of statical machine translation.
  
  \item[Chapter \ref{chapter:extractors}: Online Grammar Extractors] \hfill \vspace{0.25cm} \\    
	The traditional approach for storing translation models in memory is achieved with the help of phrase tables, dictionary-like data structures that map all source phrases from the training corpus to their target side correspondents. Phrase tables often become unmanageably large, especially in the case of hierarchical phrase-based systems, which make use of translation rules containing gaps. Online grammar extractors avoid loading all translation rules into memory by constructing memory efficient data structures on top of the source side of the parallel data, which are used to efficiently locate phrases in the corpus and extract translation rules on the fly during decoding. The online grammar extractor presented in this chapter extends \newcite{Lopez2007} and introduces a new technique for matching phrases containing gaps that significantly reduces the extraction time for hierarchical phrase-based systems. This chapter is based on the following publication:
  
  \begin{list}{bibrandom1}{}
    \bibitem[x1]{x1}
      Paul Baltescu and Phil Blunsom.
      \newblock 2014.
      \newblock {A Fast and Simple Online Synchronous Context Free Grammar Extractor}.
      \newblock {\em Prague Bulletin of Mathematical Linguistics}.
    \end{list}
    \nocite{Baltescu2014extractor}

  \item[Chapter \ref{chapter:nlms}: Neural Language modeling for Machine Translation] \hfill \vspace{0.25cm} \\
	In this chapter, we explore neural language models as a memory efficient alternative to traditional n-gram models. Recent research has shown positive results when neural language models are integrated as additional features in a decoder \cite{Vaswani2013,Botha2014} or when used for n-best list rescoring \cite{Schwenk2010}. These publications follow different approaches for scaling neural language models and one goal of this chapter is to conduct a thorough analysis to understand which of these techniques is best in a practical setup. We show that when memory is limited, neural language models clearly outperform traditional n-gram models, but this is not true when the memory constraint is removed. Finally, we explore extensions to neural language models with the goal of improving translation quality. The work presented in this chapter is based on the following publications:
        
    \begin{list}{bibrandom2}{}
    \bibitem[x2]{x2}
      Paul Baltescu, Phil Blunsom and Hieu Hoang.
      \newblock 2014.
      \newblock {OxLM: A Neural Language modeling Framework for Machine Translation}.
      \newblock {\em Prague Bulletin of Mathematical Linguistics}.
    \bibitem[x3]{x3}
      Paul Baltescu and Phil Blunsom.
      \newblock 2015.
      \newblock {Pragmatic Neural Language modeling in Machine Translation}.
      \newblock {\em Proceedings of the 2015 Conference of the North American Chapter of the Association for Computational Linguistics}.
    \end{list}
    \nocite{Baltescu2014oxlm,Baltescu2015}

  \item[Chapter \ref{chapter:conclusions}: Conclusions] \hfill \vspace{0.25cm} \\
  In the previous chapters, we have discussed online grammar extractors and neural language models as memory efficient alternatives to phrase tables and traditional n-gram language models. We now show that by putting these components together we obtain a scalable translation system that can operate in a memory constrained environment. We conclude our thesis by reviewing our findings and discussing avenues for future work.

\end{description}

\chapter{Statistical Machine Translation}\label{chapter:smt}

\section{Introduction}

In this chapter, we discuss the core components of a standard machine translation system. We take a pragmatic approach and go step by step through the stages involved in building a translation system from scratch, starting with large amounts of parallel and monolingual text. Our presentation closely follows the steps for building a baseline system with cdec \cite{Dyer2010} and Moses \cite{Koehn2007}, two popular open source translation toolkits. Relating to these tools is also useful considering that the work presented in the latter chapters of this thesis is tightly integrated with these frameworks. In fact, the reader can take the key components discussed here and the extensions from \autoref{chapter:extractors} and \autoref{chapter:nlms} and construct an efficient, high quality system that can work on a memory constrained device. We show this in \autoref{chapter:conclusions}.

This chapter is not a comprehensive review of techniques and trends employed in statistical machine translation. Instead, we aim to provide the minimal information that is sufficient to understand how translation systems work. We also focus on the parts relevant for laying the foundations for the following chapters. For a detailed literature review of statistical machine translation, we defer the reader to \newcite{Lopez2008b}.

In order to train a translation system, one needs large quantities of parallel data. A parallel corpus is a set of pairs of sentences that are translations of each other. Parallel corpora can be obtained from a number of sources: parliamentary proceedings \cite{Koehn2005}, news articles from international agencies, weather forecasts, instruction manuals, etc. In order for the system to produce fluent translations, one also needs large amounts of monolingual text in the target language. Monolingual data is cheaper to obtain (e.g. by crawling the web) and larger quantities are usually available. For the most common pairs of languages, one can find enough parallel and monolingual data to train a translation system on the website of the Workshop in Statistical Machine Translation. \footnote{The workshop is organized on an yearly basis. The website for the 2015 edition is \url{http://statmt.org/wmt15/}, and from there one can access previous editions.}

The first step for training a translation system is cleaning the training data. This step usually consists of tokenizing and lowercasing/truecasing the data. Tokenization is the process of breaking up sentences into individual tokens (words). The main challenge for tokenization is dealing with punctuation, i.e. good tokenizers separate out words from punctuation, but keep abbreviations (\textit{Msc., a.m.}), compound words (\textit{apr\`{e}s-midi}), etc. as single tokens. Lowercasing refers to the process of replacing all the capital letters in a word with lowercase letters. Truecasing allows each letter to be either lowercased or uppercased in order to obtain the most likely base form of a token. Lowercasing is the easier approach because of its deterministic nature. On the other hand, truecasing is more powerful because it has the ability to distinguish between words with the same lowercase form (e.g. \textit{US, us}). These techniques are applied in order to make the data statistics less sparse and the overall system more robust. Common practice dictates that pairs of very long sentences or having unusual length ratios are removed from the parallel data. These sentence pairs are frequently the outcome of flaws in the process of collecting the training data. For certain languages, additional text processing may be needed (e.g. word segmentation for Chinese).

The next step is aligning the parallel corpus. Alignment models are discussed in further detail in \autoref{section:smt:alignment_models}. The parallel corpus and the word alignments are used to extract translation rules. The structure of translation rules depends on the formalism chosen as foundation for the translation system. In \autoref{section:smt:translation_formalisms}, we discuss finite state transducers and synchronous context free grammars as underlying formalisms for translation systems. We also discuss phrase tables as the default approach to storing translation rules in memory. \autoref{section:smt:language_models} reviews the use of language models in machine translation and discusses back-off n-gram models as the standard approach for language modeling. \autoref{section:smt:scoring_model} shows how translation rules, language models and other signals are combined together as part of a unified scoring model. \autoref{section:smt:decoding} explains the algorithms for translating a source sentence into a target sentence, a process also known as \textit{decoding}. \autoref{section:smt:evaluation} concludes by discussing methods for evaluating the quality of translation systems.

\section{Alignment Models}\label{section:smt:alignment_models}

Parallel corpora are an essential resource for training translation models. However, they cannot be used for this task in their raw form, because they cover an insignificant part of the space of sentence pairs. The observed sentence pairs are  also unlikely to repeat again. Therefore, we need to use the parallel data to learn smaller translation units which can later be composed in order to translate full sentences. Alignment models help with this crucial task by learning the process of translating (or aligning) single words. Larger translation units can then be obtained by composing several aligned words located in the same context (\autoref{section:smt:translation_formalisms}).

\begin{figure}
\begin{center}
\includegraphics[scale=0.4]{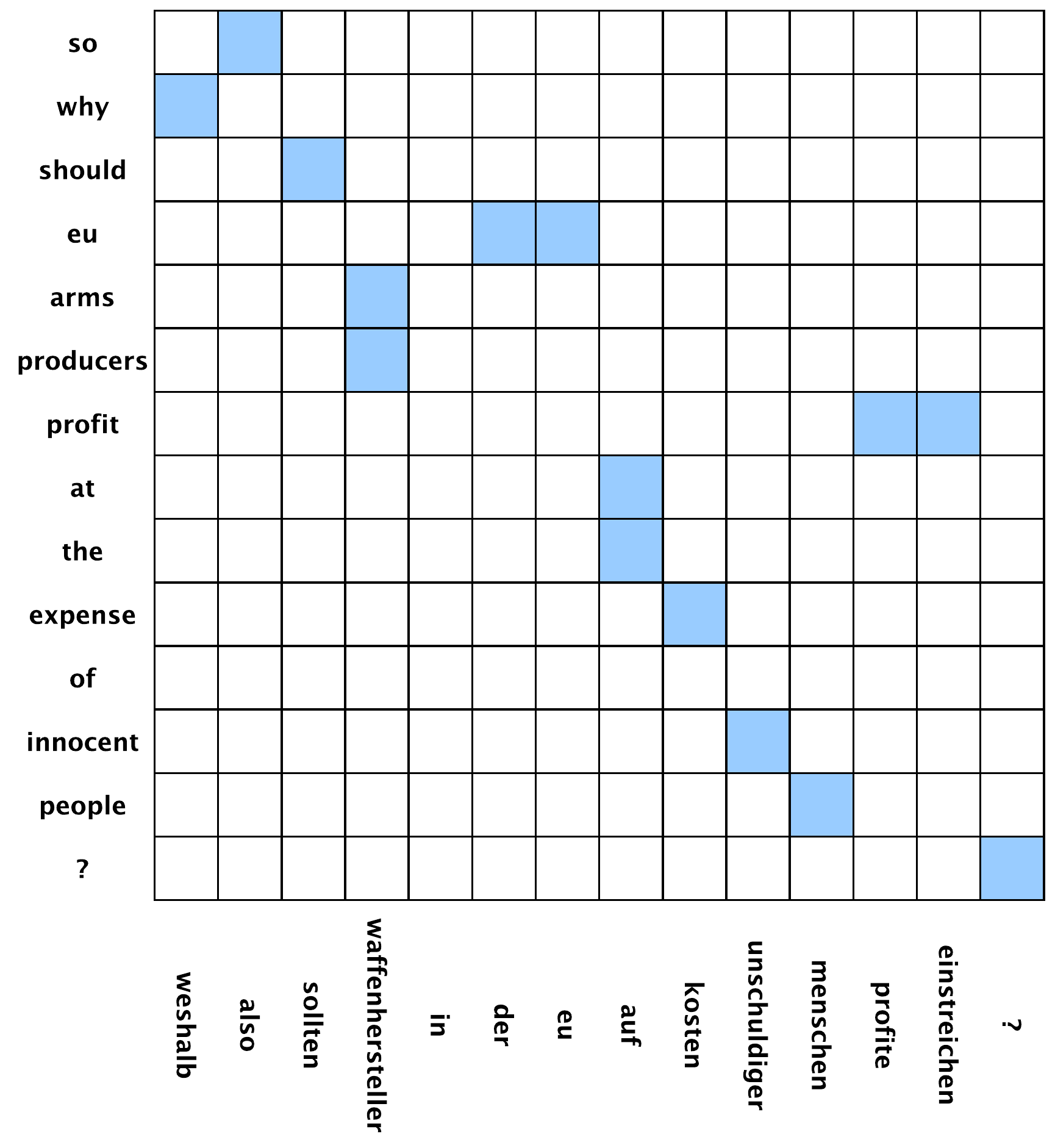}
\end{center}
\caption[Alignment matrix]{Alignment matrices are a visual interpretation of word alignments for a pair of sentences. In this image, one can notice that the alignment links follow the general direction of the main diagonal, a common trait for Indo-European languages. The alignment matrix also illustrates different syntactic rules in the language pair. In this example, one can observe that English has subject-verb-object (SVO) syntax, while in German, the more frequent SOV structure is present.}
\label{figure:alignment_matrix}
\end{figure}

Formally, let $\textbf{s} = s_1^N$ and $\textbf{t} = t_1^M$ be a pair of sentences from the parallel corpus. An alignment $\textbf{a}$ is a subset of index pairs from $[1, N] \times [1, M]$. We say $(i, j) \in \textbf{a}$ if the words $s_i$ and $t_j$ are translations of each other. Word alignments can be illustrated graphically using an alignment matrix as shown in \autoref{figure:alignment_matrix}.

\newcite{Brown1993} made the first substantial leap in statistical machine translation by introducing several word level translation models (known as the IBM Models), which are now regarded as the standard set of models for word alignment. \newcite{Vogel1996} used Hidden Markov Models to define an alignment model known as the HMM alignment model. By default, the Moses toolkit \cite{Koehn2007} relies on Giza++ \cite{Och2003} for aligning the parallel corpus. Giza++ is a fine tuned implementation of the IBM alignment models, using the first models to initialize the more advanced ones. In Giza++, the IBM Model 2 is replaced by the HMM alignment model. The cdec toolkit computes alignments using an adaptation of IBM Model 2 with fewer parameters \cite{Dyer2013}.

For demonstration purposes, let us analyze the HMM alignment model into greater detail. The HMM model is an \textit{asymmetric} alignment model (and so are the IBM Models) because each target word is aligned to exactly one source word, while a source word can be aligned to multiple target words. A dummy null token is introduced on the source side to permit unaligned target words. The model encapsulates a bias for monotonic alignments, i.e. consecutive target words are more likely to be aligned to consecutive source words. Formally, the model defines the joint probability of a target sentence $\textbf{t}$ and an alignment $\textbf{a}$ given a source sentence $\textbf{s}$ as follows:
\begin{equation}
p(\textbf{t}, \textbf{a} | \textbf{s}) = \prod_{j=1}^M p(a_j | a_{j-1}, N) p(t_j | s_{a_j})
\end{equation}
The model is defined in terms of the translation probabilities $p(t | s)$ and the alignment penalties $p(i | j, N)$. If the word alignments were known ahead of time, computing these parameters would be straightforward. Conversely, if we knew the translation probabilities and alignment penalties, we could find the optimal alignment using the Viterbi algorithm \cite{Viterbi1967}. However, since neither are known ahead of time, we must resort to the EM algorithm for parameter estimation \cite{Baum1972}. The other alignment models follow the same general approach with regards to learning the parameters and finding the maximum probability alignment.

The standard practice for obtaining symmetric alignments is to apply the alignment models in both directions and to use some heuristic to combine the resulting alignments (e.g., set union, set intersection, etc.). The default heuristic used by the Moses and cdec toolkits starts from the intersection of the two alignments and then adds additional links along the main diagonal of the alignment matrix \cite{Koehn2010}. The heuristic includes a final step adding links from the set union to those words that remain unaligned.

\section{Translation Models}\label{section:smt:translation_formalisms}

Translation models define the set of translation rules employed by a translation system. Most translation models stem from one of the following two formalisms: finite state transducers (FSTs) or synchronous context free grammars (SCFGs). These formalisms are similar in nature to their better known monolingual counterparts, the finite state automata and the context free grammars, but have the ability to model a target language in addition to the source language, which makes them suitable for machine translation. In this section, we present their formal definitions and briefly describe one model of each type. We also introduce phrase tables, the standard approach for storing translation models in memory, and explain how they are constructed from a word aligned parallel corpus.

\subsection{Finite State Transducers}\label{subsection:fst}

Formally, a FST is a tuple $(Q, V_s, V_t, D)$, where $Q$ represents a set of states, $V_s$ and $V_t$ are sets of symbols and $D \subseteq Q \times V_s \times V_t \times Q$ is a set of transitions. In the context of machine translation, $V_s$ and $V_t$ define the set of source and target tokens (words, phrases, etc.), the states $Q$ are a succinct representation of translation hypotheses, and the transitions $D$ specify the set of base rules that define the space of valid translations. For example, transitions can model the process of translating single word units as follows:  a transition $q_1 \xrightarrow{s/t} q_2$ with $q_1, q_2 \in Q, s \in V_s, t \in V_t$ represents the source word $s$ being translated to the target word $t$, extending the current translation hypothesis represented by $q_1$ into $q_2$. The states $Q$ also indicate how close we are to obtain a full translation of the input sentence. In practice, most FST-based translation systems are described as compositions of FSTs. The resulting systems are also FSTs because FSTs are closed to the composition operator.

Historically, the first translation models to show promising results were based on the FST formalism and consequently these models laid the foundations of statistical machine translation. Examples of FST-based models include the IBM Models \cite{Brown1993}, the HMM alignment model \cite{Vogel1996} and the phrase-based translation models \cite{Koehn2003}. We briefly touch upon phrase-based models as described in \newcite{Koehn2003} because these models were employed in many state of the art systems and are still widely in use today. In fact, the Moses toolkit \cite{Koehn2007} relies by default on phrase-based models. Compared to their word-based predecessors, phrase-based models can translate contiguous groups of words (phrases) in a single step. The intuition behind these models is that if a phrase pair is observed enough times in the parallel corpus, it is more likely to produce the intended translation than if we replace individual source words with their most likely translations. Another reason why phrase-based models perform better is their ability to capture local reorderings and to improve grammatical agreement.

\begin{figure}
\begin{center}
\includegraphics[scale=0.5]{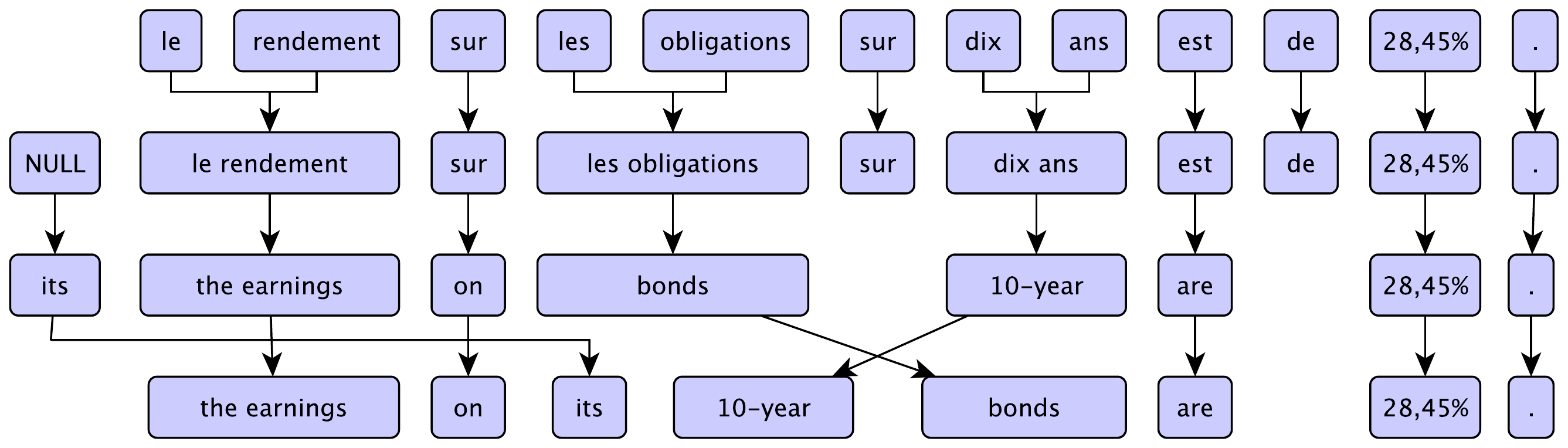}
\end{center}
\caption[Translating with phrase-based models]{The three step process for translating with phrase-based models.}
\label{figure:translation_process}
\end{figure}

At a high level, phrase-based translation can be seen as a three steps process: first, the source sentence is segmented into several phrases, then each source phrase is translated into a target phrase, and finally, these phrases are reordered to produce a fluent translation in the target language (\autoref{figure:translation_process}). Each of these steps can be modelled using an FST and thus phrase-based translation models are a composition of FSTs. The first two tasks are straightforward to solve with an FST, but FSTs are not well suited to deal with reorderings. In fact, an FST must have $O(2^N)$ states in order to capture all the $O(N!)$ permutations of a set of $N$ words. Furthermore, \newcite{Knight1999} showed that the problem of finding the optimal reordering is equivalent to the traveling salesman problem and thus NP-complete. As a result, decoding with phrase-based models cannot be done efficiently and one must resort to heuristics like beam search \cite{Koehn2004} (\autoref{subsection:fst_decoding}). Reordering poses the same challenge to all FST-based models.

\subsection{Synchronous Context Free Grammars}\label{subsection:scfg}

SCFG-based models were introduced with the goal of alleviating some of the problems with FST models. First, SCFG models can capture long distance reorderings more easily than FST models. Second, they provide a framework for learning discontiguous phrases, allowing the models to learn useful translation templates, e.g. \textit{he is X years old} $\longleftrightarrow$ \textit{il a X ans}. Finally, SCFGs provide support for bridging the gap between translation and syntax.

\begin{figure}
\begin{subfigure}{.35\textwidth}
\small
\vspace{2cm}
(1) NP $\rightarrow$ NP$_1$ sur NP$_2$ $|$ NP$_2$ NP$_1$ \\
(2) NP $\rightarrow$ les NN$_1$ $|$ NN$_1$ \\
(3) NP $\rightarrow$ NN$_1$ NN$_2$ $|$ NN$_1$ NN$_2$ \\
(4) NN $\rightarrow$ obligations $|$ bonds \\
(5) NN $\rightarrow$ dix $|$ 10 \\
(6) NN $\rightarrow$ ans $|$ year \\
\end{subfigure}
\begin{subfigure}{.65\textwidth}
\begin{center}
\includegraphics[scale=0.5]{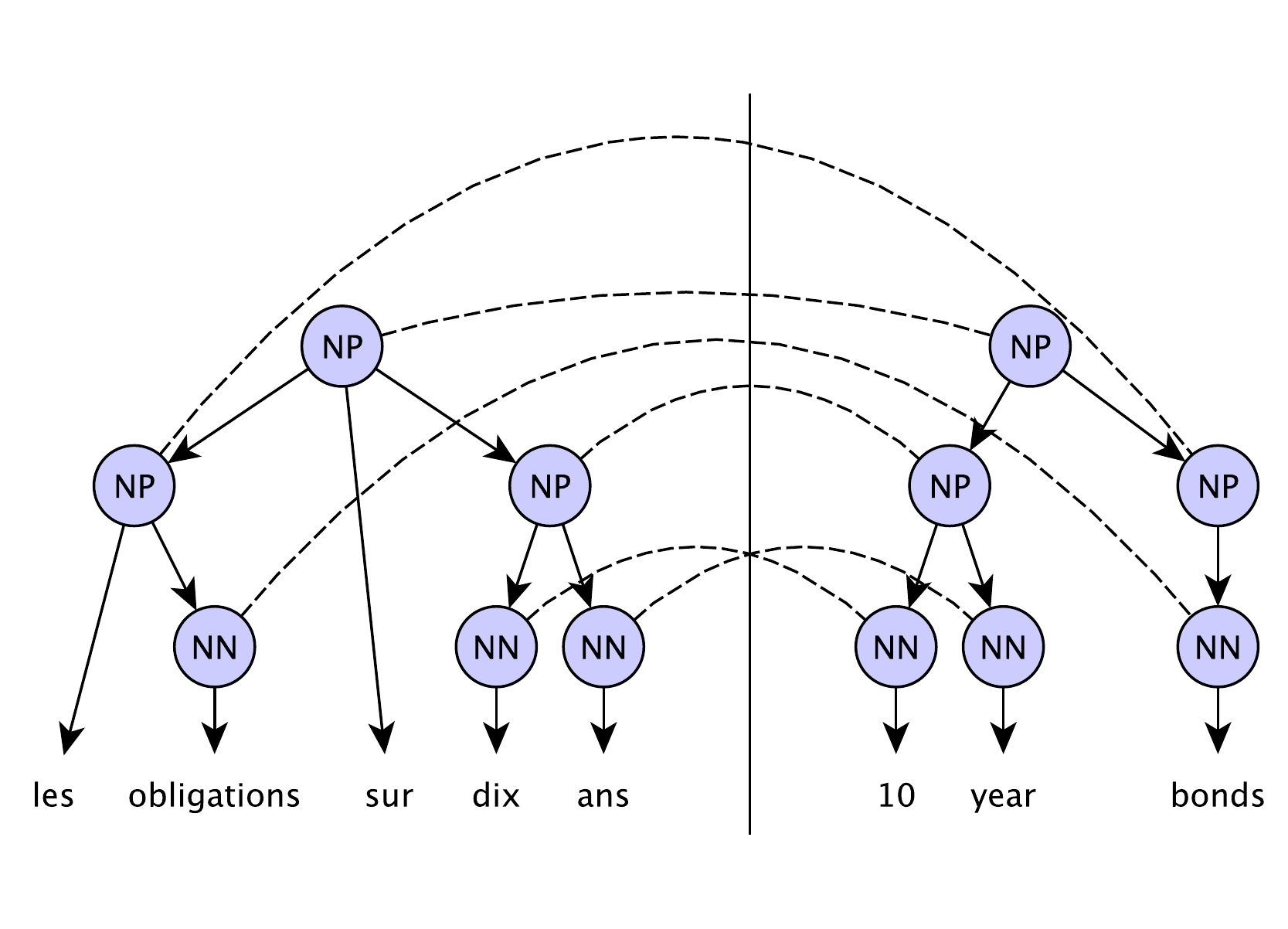}
\end{center}
\end{subfigure}
\caption{A simple SCFG grammar and an example derivation.}
\label{figure:scfg}
\end{figure}

Synchronous context free grammars are an extension of context free grammars capable of modeling an additional language. They are formally defined as a tuple $(N, S, V_s, V_t, D)$, where $N$ is a set of nonterminals, $S \in N$ is the starting nonterminal for any SCFG derivation, $V_s$ is a set of source terminals, $V_t$ is a set of target terminals and $D \subseteq N \times (N \times V_s)^{*} \times (N \times V_t)^{*}$ is a set of productions (rules). A production of the form $X \rightarrow \alpha\ |\ \beta$ indicates that a nonterminal $X \in N$ may be rewritten as a string of terminals and nonterminals $\alpha \in (N \times V_s)^{*}$ in the source language and $\beta \in (N \times V_t)^{*}$ in the target language. The source and target side of a rule must contain the exact same multiset of nonterminals and each source nonterminal must be aligned to exactly one target nonterminal of the same type. Nonterminals on the right hand side of a rule are usually labelled with numbers to indicate how they align to each other. \autoref{figure:scfg} shows an example SCFG and illustrates how this grammar can be used for translating sentences.

Modeling word reordering with an SCFG is a trivial task as one can simply enumerate rules for every possible permutation up to some given length. However, decoding algorithms for models based on this formalism are adaptations of the well known parsing algorithms for CFGs (\autoref{subsection:scfg_decoding}) and therefore their complexity depends on the size of the grammar. Adding all these permutations to the grammar would result in an exponential number of rules yielding impractical decoding algorithms. A common compromise is to include only the following two reordering rules: $X \rightarrow X_1 X_2\ |\ X_1 X_2$ and $X \rightarrow X_1 X_2\ |\ X_2 X_1$. A number of other permutations may be obtained by repeatedly applying these two rules, but not all (e.g. $X \rightarrow X_1 X_2 X_3 X_4\ |\ X_2 X_4 X_1 X_3$). Although the number of permutations that cannot be obtained with this procedure increases exponentially with the size of the permutation \cite{Wu1997}, \newcite{Zens2003} show that only a negligible part of real world reorderings cannot be captured with this simplified model.

A number of translation models use the SCFG formalism: inversion transduction grammars \cite{Wu1997}, hierarchical phrase-based models \cite{Chiang2007}, syntactic tree-to-string grammars \cite{Galley2004,Galley2006}, etc. We briefly describe hierarchical phrase-based models because of their relative success over phrase-based models. The cdec toolkit \cite{Dyer2010} relies by default on a hierarchical phrase-based model.

Hierarchical phrase-based models are SCFGs with a single nonterminal (usually labeled as $X$). The source and target vocabularies directly map to the vocabularies of the language pair for which the model is built. To allow for efficient decoding, hierarchical phrase-based models enforce the restriction discussed previously and limit the number of nonterminal pairs in the right hand side of a rule to 2. Decoding with hierarchical phrase-based models is polynomial in the size of the sentence, but the main challenge is incorporating the language model. We cover this subject in greater detail in \autoref{subsection:scfg_decoding}.

\subsection{Phrase Tables}\label{subsection:phrase_tables}

In \autoref{subsection:fst} and \autoref{subsection:scfg}, we presented the two most common formalisms that define the structure of the rules employed by translation systems. In this subsection, we focus on how these rules are stored in memory and explain how they are extracted from a word-aligned parallel corpus (a task known as phrase or grammar extraction).

The traditional approach to storing translation rules in memory is achieved via phrase tables. A phrase table maps each source phrase to all the target phrases that align to it in the parallel text. For each target phrase, the phrase table also stores an additional set of scores whose role is discussed in \autoref{section:smt:scoring_model}. Phrase tables are often pruned so only the most frequent phrase pairs are kept. Pruning can be done based on the weights associated with each target phrase, or by keeping a limited number of target phrases for each source phrase. Decoders require the ability to efficiently look up the target phrases associated with any source phrase. As a result, phrase tables are usually implemented as hash tables or tries to allow constant time lookups.

Phrase tables are constructed offline in a preprocessing step taking as input the parallel corpus and the word alignments. The algorithm processes the corpus sentence by sentence and extracts all valid phrase pairs. A phrase pair is considered valid if none of its source or target words aligns to a word outside of the phrase pair. For FST rules, the extraction algorithm \cite{Och2004} iterates through all the source phrases $s_i^j$ of a given sentence $s_1^N$,  $1 \leq i \leq j \leq N$. For every word $s_k$, $i \leq k \leq j$, all the target words aligned with $s_k$ are added to a set $T$. After the whole source phrase is processed, the algorithm checks if the target words in $T$ form a contiguous phrase in the target sentence and that none of the target words are aligned with source words outside $s_i^j$. Additional phrase pairs will be extracted if there are unaligned target words adjacent to the target phrase defined by $T$. The extraction algorithm for SCFG rules works in a similar fashion, but it contains a large number of additional edge cases because it must ensure that gaps are also correctly aligned. A detailed account of the extraction algorithm for SCFG rules is given in \newcite{Lopez2008}.

Parallel corpora used for training translation systems usually contain millions of sentences. As a result, phrase tables easily grow too large to be held in memory. For FST models, the number of contiguous subphrases of a sentence of length $N$ is $O(N^2)$. A common solution is to limit the length of the extracted phrases by a threshold $L$, reducing the number of subphrases to $O(N \times L)$. This is a sensible restriction because longer phrase patterns are less likely to be observed during decoding. For example, \newcite{Koehn2003} recommend extracting phrase pairs having up to four words on the source side.

For SCFG-based models, the number of phrases that can be extracted from a single sentence is exponential in the sentence length. This number can be significantly reduced when a threshold is placed on the maximum number of nonterminal pairs in a rule (like in hierarchical phrase-based models) and when the maximum span of a rule is limited (similar to FST models). However, even with these limitations in place, discontiguous translation rules continue to be orders of magnitude more than contiguous rules. Traditional phrase tables are not suitable for SCFG grammar extraction at scale, even without taking into account additional memory constraints, such as the ones imposed by mobile devices. A compact and scalable alternative to phrase tables is presented in \autoref{chapter:extractors}.

\section{Language Models}\label{section:smt:language_models}

The language model is a key component in a translation system which is responsible for the fluency of the output translations. It drives a good part of the end to end translation quality and  considerable improvements can often be achieved by using more monolingual data to train the language model. In this section, we explain how language models work, while \autoref{section:smt:scoring_model} shows how translation models and language models are combined together in a unified scoring model.

Language models are statistical models used to score how likely a sequence of words is to occur in a certain language by means of a probability distribution. Let \textbf{w} = $w_1^M$ be a sentence in the target language and $P(\textbf{w})$ the probability distribution defined by the model. According to the chain rule of probability, $P(\textbf{w})$ can be decomposed as the product of the probabilities of each target word given its preceding context:
\begin{equation}
P(\textbf{w}) = \prod_{i=1}^M P(w_i | w_1^{i-1}).
\end{equation}
To prevent the model from relying on distributions computed from very sparse statistics, a $n$-1th order Markov assumption is typically incorporated in the model:
\begin{equation}
P(\textbf{w}) = \prod_{i=1}^M P(w_i | w_{i-n+1}^{i-1}).
\end{equation}

Back-off n-gram models are the default language modeling implementation used in machine translation. These models estimate the conditional probability as:
\begin{equation}
P(w_i | w_{i-n+1}^{i-1}) = \frac{c(w_{i-n+1}^i)}{c(w_{i-n+1}^{i-1})},
\end{equation}
where $c(w_{i-n+1}^i)$ and $c(w_{i-n+1}^{i-1})$ represent the number of times $w_{i-n+1}^i$ and $w_{i-n+1}^{i-1}$ are observed in the monolingual corpus. In their raw form, n-gram language models do not accurately estimate rare n-grams. Over the years, a number of smoothing techniques have been proposed in order to address this problem \cite{Jelinek1980,Katz1987,Kneser1995,Chen1999}.

Machine translation toolkits like Moses \cite{Koehn2007} and cdec \cite{Dyer2010} have plugins for several open source implementations of back-off n-gram models:  SRILM \cite{Stolcke2002}, IRSTLM \cite{Federico2008} and KenLM \cite{Heafield2011}. \newcite{Heafield2011} shows his implementation is superior to SRILM and IRSTLM both in terms of speed and memory usage. In fact, two separate implementations are provided as part of KenLM. One is optimized for speed and uses a hash table with linear probing to look up n-gram weights. The other, optimized for memory, but still faster than the alternatives, relies on a trie and makes use of floating point quantization.

Back-off n-gram models are used as the default language modeling choice in machine translation because they produce very good results if enough monolingual data is available. They are also fast to train and query and their definition is very intuitive. On the other hand, a n-gram language model stores a numerical value for every n-gram in the training corpus. As a result, even the most compact implementations (e.g. the KenLM trie implementation) require tens of gigabytes of memory for a decently sized monolingual corpus (see \autoref{section:nlms:comparison}). In conclusion, back-off n-gram models are not suitable for memory constrained environments. \autoref{chapter:nlms} investigates neural language models as a space-efficient alternative to n-gram language models.

\section{Scoring Model}\label{section:smt:scoring_model}

The translation formalisms introduced in \autoref{section:smt:translation_formalisms} consist of a set of rules which define the entire set of valid translations. However, not all of these translations are equally good. The scoring model provides a framework for comparing these translations by assigning a probability to all output sentences. The decoder's goal (\autoref{section:smt:decoding}) is to infer the best translation with respect to the scoring model.

The goal of a translation system is to find $\argmax_\textbf{t} P(\textbf{t} | \textbf{s})$, where $\textbf{s}$ is the source sentence and $\textbf{t}$ is a possible translation. It is useful to extend this definition to include the translation rules used by the system in order to produce $\textbf{t}$. Let $\textbf{d}$ denote a set of rules (derivation) and $\mathcal{S}(\textbf{t})$ be the set of derivations $\textbf{d}$ which produce $\textbf{t}$. We can rewrite our objective as $\argmax_\textbf{t} \sum_{\textbf{d} \in \mathcal{S}(\textbf{t})} P(\textbf{t}, \textbf{d} | \textbf{s})$. Unfortunately, this optimization problem is intractable for both FSTs and SCFGs. The common practice is to use the Viterbi approximation instead: $\argmax_{\textbf{t}, \textbf{d} \in \mathcal{S}(\textbf{t})} P(\textbf{t}, \textbf{d} | \textbf{s})$.

Most translation systems can trace their roots to the IBM Models \cite{Brown1993}. \newcite{Brown1993} represent the translation process as a generative model:
\begin{equation}
P(\textbf{t}, \textbf{d} | \textbf{s}) = \frac{P(\textbf{s}, \textbf{t}, \textbf{d})}{P(\textbf{s})} = \frac{P(\textbf{t}) P(\textbf{s}, \textbf{d} | \textbf{t})}{P(\textbf{s})} \propto P(\textbf{t}) P(\textbf{s}, \textbf{d} | \textbf{t}).
\end{equation}
Here,  $P(\textbf{s}, \textbf{d} | \textbf{t})$ is a target-to-source translation model (the reverse of the models discussed in \autoref{section:smt:translation_formalisms}) and $P(\textbf{t})$ is the target language model (\autoref{section:smt:language_models}). By combining these two terms we aim to obtain a translation that is both accurate and fluent in the target language.

The language model is traditionally estimated as described in \autoref{section:smt:language_models}. The translation model probabilities are estimated from a word aligned parallel corpus, also via frequency counts. For example, to compute the probability that the word \textit{president} translates as \textit{pr\'{e}sident}, we divide the number of times the two words are aligned to each other in a French-English parallel corpus with the number of times the word \textit{president} appears in the target side of the corpus. For phrases (contiguous or not), we can multiply the word level translation probabilities (averaging the probabilities for words with several alignments) or we can use phrase level frequency counts instead. The latter approach is usually more accurate. These weights are stored in the phrase table in addition to each target phrase.

The problem with generative translation models is that they need to make too many independence assumptions in order to be tractable. For example, when applying a translation rule, one can only use the source words belonging to the current phrase pair as signal. Other information is ignored; in this case, the entire source sentence can be useful when choosing the target words. The solution for this problem is to rely on discriminative models which permit the use of overlapping features. Indeed, most translation systems today, including Moses \cite{Koehn2007} and cdec \cite{Dyer2010}, use a discriminative model for scoring derivations.

The most common form of discriminative translation models is log-linear models. A log-linear model defines a set of $K$ feature functions $f_1^K(\textbf{s}, \textbf{t}, \textbf{d})$ producing strictly positive outputs and learns a set of weights $\lambda_1^K$. The weights $\lambda_1^K$ capture the correlation between the feature functions and the model's output: a large positive $\lambda_k$ implies the function $f_k$ is a strong predictor of the model's output, a large negative $\lambda_k$ shows a strong inverse correlation between $f_k$ and the model's predictions, while a $\lambda_k$ close to 0 shows that $f_k$ is not useful for predictions. The feature functions $f_1^K$ typically include a target language model and a generative target-to-source translation model, just like generative scoring models. However, since log-linear models support overlapping features, it is common to also include a source-to-target generative translation model, to use several scoring techniques for translation models (e.g. word level, phrase level, word embeddings) or to include multiple target language models. Using all these signals, a log-linear model defines a probability distribution as follows:
\begin{equation}
P(\textbf{t}, \textbf{d} | \textbf{s}) = \frac{\exp \sum_{k=1}^K \lambda_k f_k(\textbf{s}, \textbf{t}, \textbf{d})}{\sum_{\textbf{t}', \textbf{d}' \in \mathcal{S}(\textbf{t}')} \exp\sum_{k=1}^K \lambda_k f_k(\textbf{s}, \textbf{t}', \textbf{d}')} .
\end{equation} 
Fortunately, most decoding algorithms do not require a well-defined probabilistic model, so the intractable normalization term can be ignored.

The first step towards training a log-linear translation model is computing the values of the generative features as explained above. Once these features are known (or can be computed online quickly), we can proceed with learning the coefficients $\lambda_1^K$. One option is to use maximum likelihood estimates learned via gradient descent on a small development corpus. Unfortunately, this method requires computing the full normalization term. The standard optimization applied in this case is to approximate the normalization term with the list of n-best translations, hoping that they account for most of the probability mass.

\autoref{section:smt:evaluation} discusses how the quality of machine translation systems is evaluated. \newcite{Och2003b} shows that optimizing the weights of the scoring model directly towards the evaluation metric results in considerable qualitative gains. The key difficulty is that these metrics are non-differentiable and require new training algorithms. The default algorithm for training translation models with the Moses toolkit is minimum error rate training (MERT) \cite{Och2003b}. MERT was also the default setting in the cdec toolkit, until recently when it was replaced with the maximum infused relaxed algorithm (MIRA) \cite{Eidelman2012}.

For demonstration purposes, let us discuss the MERT algorithm in greater detail. MERT assumes the existence of an error function $E(\textbf{\^{t}}, \textbf{t})$ defining the amount of mismatch between a translation candidate $\textbf{\^{t}}$ and the reference translation $\textbf{t}$. The goal of MERT is to find $\lambda_1^K$ which minimizes the total error on the development corpus $\mathcal{D}$:
\begin{equation}
\lambda_1^K = \argmin_{\lambda_1^K} \sum_{\textbf{s}, \textbf{t} \in \mathcal{D}} E(\argmax_{\textbf{\^{t}}} P_{\lambda_1^K}(\textbf{\^{t}} | \textbf{s}) , \textbf{t}).
\end{equation}

The algorithm has several iterations. During one iteration, we generate several candidates $\lambda_1^K$ randomly. For each candidate, we iterate over each $\lambda_k$ and try to optimize it with respect to the total error function while keeping all the other parameters $\lambda_{k' \neq k}$ constant. We keep track of the parameters $\lambda_1^K$ that minimize the error function and, if during one iteration none of the optimized candidates yield any improvement over the current solution, we terminate the algorithm.

\begin{figure}
\begin{subfigure}{.3\textwidth}
\begin{center}
\includegraphics[scale=0.2]{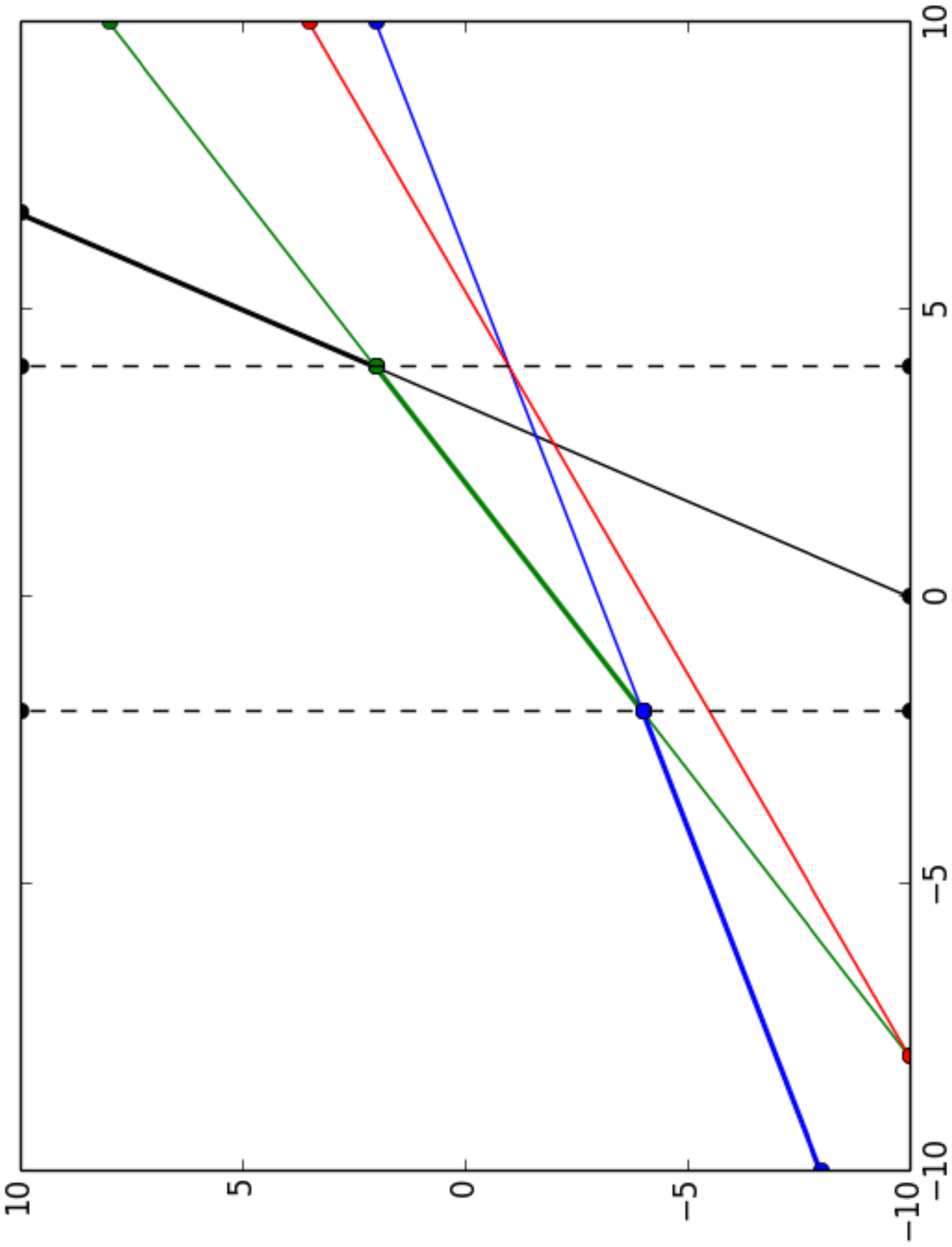}
\caption{}
\label{subfig:mert1}
\end{center}
\end{subfigure}
\begin{subfigure}{.3\textwidth}
\begin{center}
\includegraphics[scale=0.2]{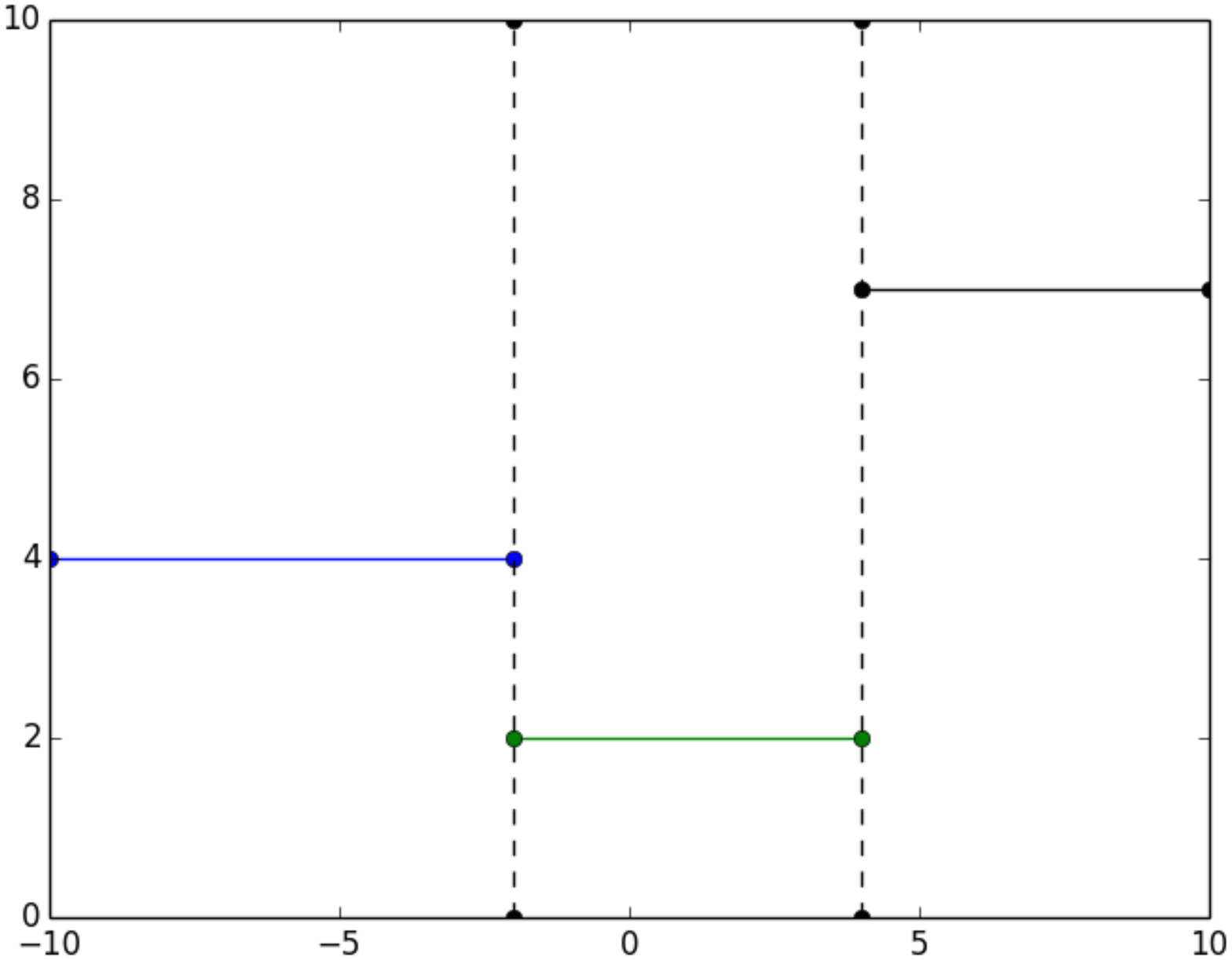}
\caption{}
\label{subfig:mert2}
\end{center}
\end{subfigure}
\begin{subfigure}{.3\textwidth}
\begin{center}
\includegraphics[scale=0.2]{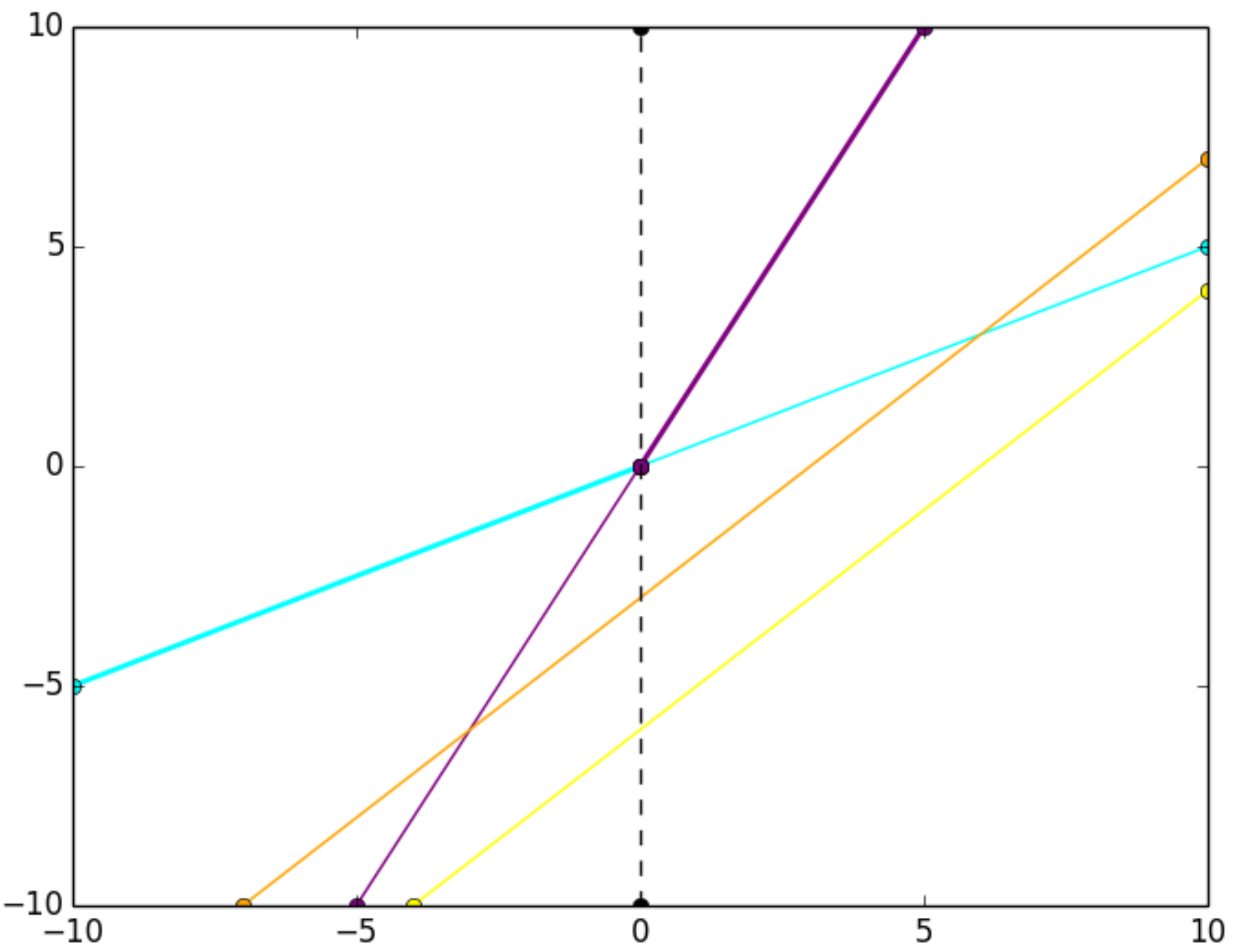}
\caption{}
\label{subfig:mert3}
\end{center}
\end{subfigure}
\begin{subfigure}{.5\textwidth}
\begin{center}
\includegraphics[scale=0.2]{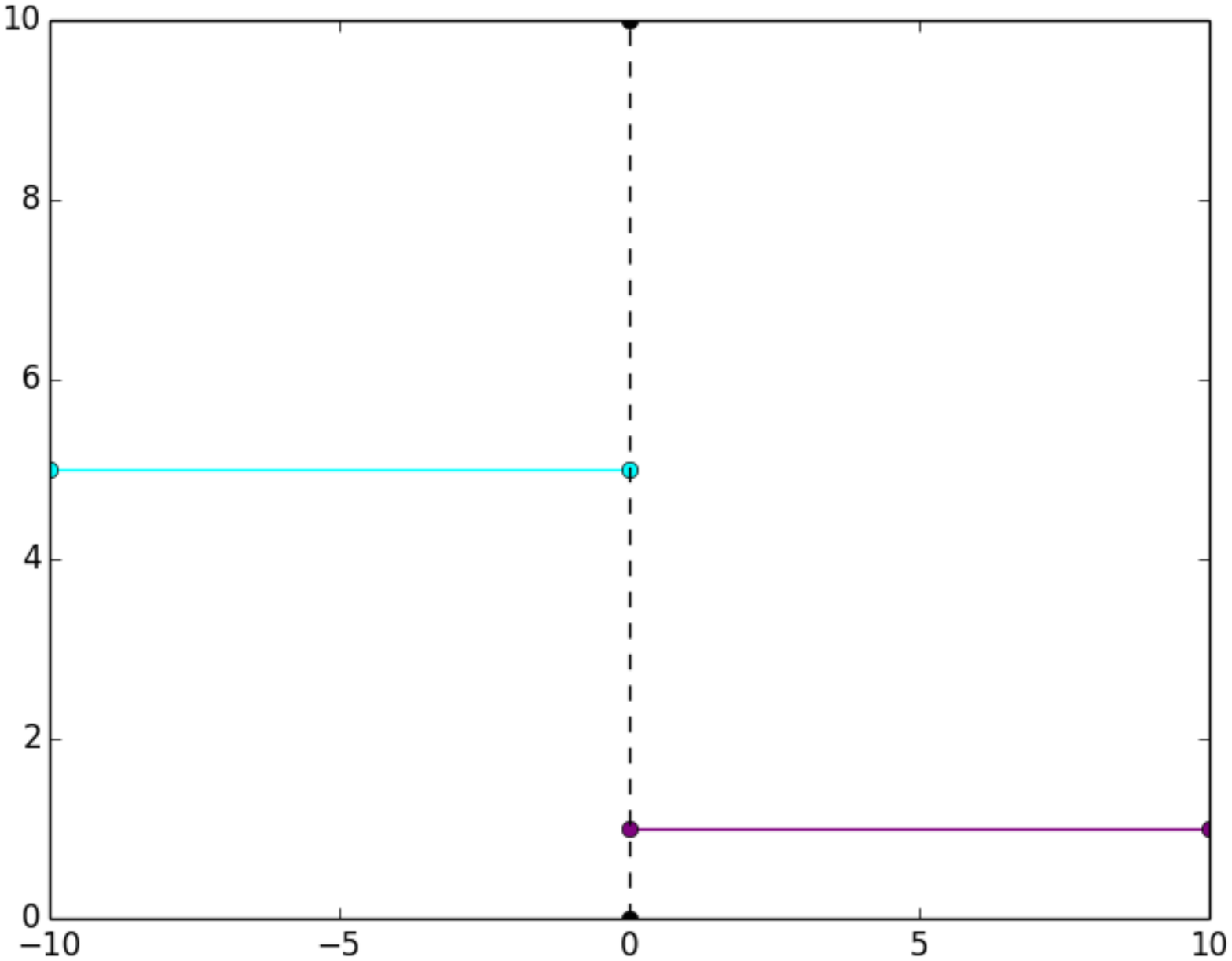}
\caption{}
\label{subfig:mert4}
\end{center}
\end{subfigure}
\begin{subfigure}{.5\textwidth}
\begin{center}
\includegraphics[scale=0.2]{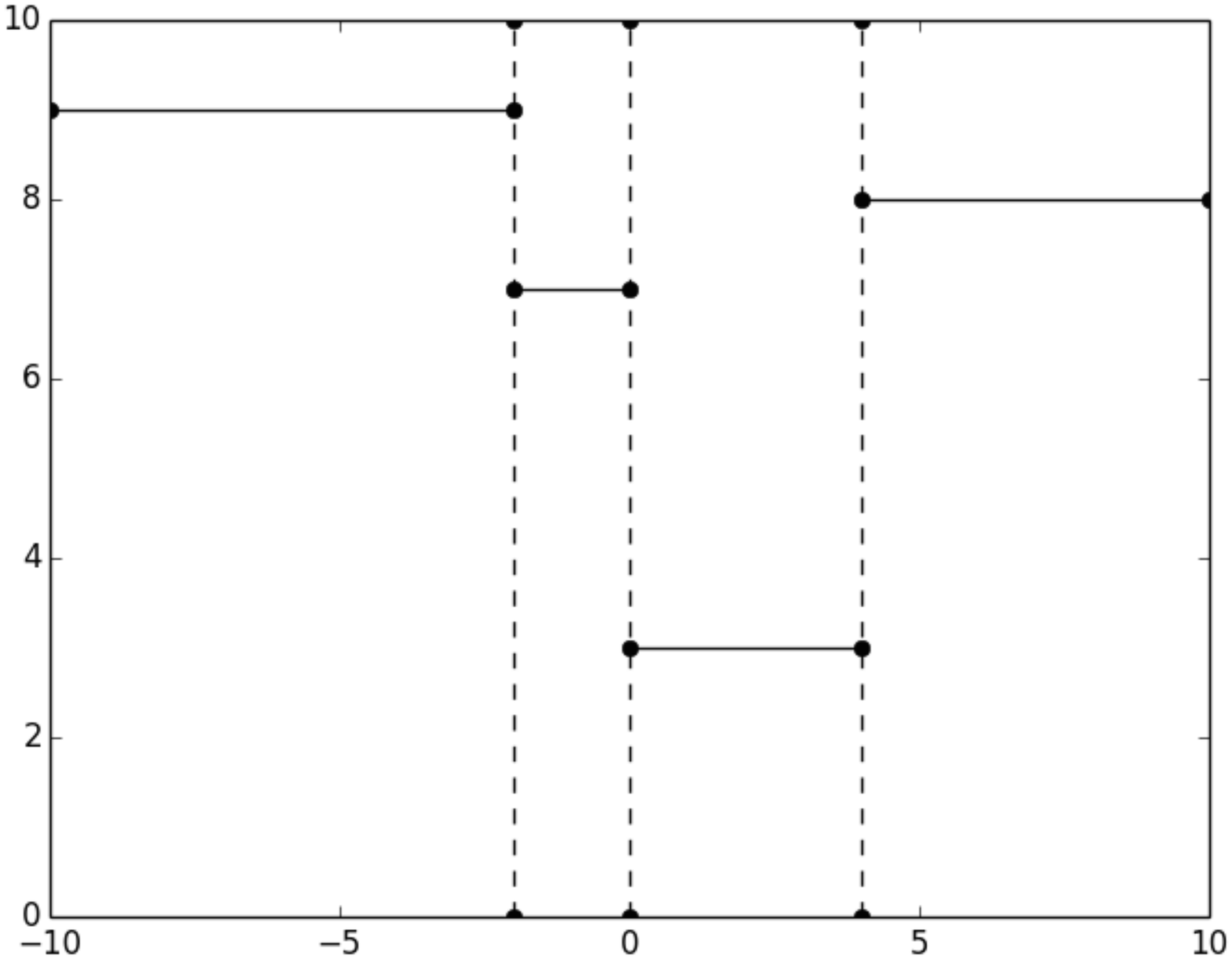}
\caption{}
\label{subfig:mert5}
\end{center}
\end{subfigure}
\caption[MERT subroutine for optimizing a single weight]{The geometric intuition behind the MERT subroutine for optimizing $\lambda_k$. \autoref{subfig:mert1} shows the lines $P(\textbf{\^{t}} | \textbf{s})$ as a function of $\lambda_k$. The interval where each $\textbf{\^{t}}$ dominates the other n-best list translations is marked by a thicker line. \autoref{subfig:mert2} shows the error function $E(\argmax_{\textbf{\^{t}}} P(\textbf{\^{t}} | \textbf{s}) , \textbf{t})$. \autoref{subfig:mert3} and \autoref{subfig:mert4} show $P(\textbf{\^{t}} | \textbf{s})$ and the error function for a different source sentence $\textbf{s}.$ \autoref{subfig:mert5} shows the total error function. For this example, any $\lambda_k$ between 0 and 4 is considered optimal; our algorithm chooses 2.} 
\end{figure}

The trick for optimizing each individual parameter $\lambda_k$ is noting that for a given translation candidate $\textbf{\^{t}}$, the function $P(\textbf{\^{t}} | \textbf{s}) = \lambda_k f_k(\textbf{s}, \textbf{\^{t}}) + \sum_{k' \neq k} \lambda_{k'} f_{k'}(\textbf{s}, \textbf{\^{t}})$ is linear in $\lambda_k$, if all $\lambda_{k' \neq k}$ are constant. For a fixed source sentence $\textbf{s}$, we take each $\textbf{\^{t}}$ belonging to the $n$-best list of $\textbf{s}$ after the previous iteration of the algorithm (as a representative subsample of the search space) and we intersect the lines defined by $P(\textbf{\^{t}} | \textbf{s})$. We obtain several compact intervals where different $\textbf{\^{t}}$ are the most likely translations of $\textbf{s}$ (\autoref{subfig:mert1}, \autoref{subfig:mert3}). For each such interval, the error function $E(\textbf{\^{t}} , \textbf{t})$ is constant because $\textbf{\^{t}}$ does not change, implying that $E(\textbf{\^{t}} , \textbf{t})$ is a step function with at most $n$ different values (\autoref{subfig:mert2}, \autoref{subfig:mert4}), $n$ being the number of candidates in the $n$-best list. By adding up the step functions for each source sentence $\textbf{s}$, we observe that the total error function is also a step function, albeit finer grained (\autoref{subfig:mert5}). $\lambda_k$ is found by iterating over the distinct values of the total error function and choosing any point (e.g. the middle point for robustness) from the interval with the smallest error.
 
\section{Decoding}\label{section:smt:decoding}

In a machine translation system, the decoder is the component responsible for translating a source sentence into a target sentence or for producing a list of $n$-best translations. At a high level, the decoder generates a set of partial translation hypotheses by repeatedly applying rules licensed by the translation model (\autoref{section:smt:translation_formalisms}) and by scoring them with the scoring model (\autoref{section:smt:scoring_model}). In this section, we review how decoding is done for both FST and SCFG-based translation models.

Both FST and SCFG models have a high degree of ambiguity and define a massively exponential space of valid translations for every source sentence. Aggressive fine-tuned optimization techniques need to be applied in order to keep decoders practical and to maintain a high bar for translation quality. This section reviews some of these techniques.

\subsection{Decoding with FST Models}\label{subsection:fst_decoding}

In order to illustrate how decoding with FST models works, we again choose phrase-based translation models (\autoref{subsection:fst}) as a representative example from this class of models. Our review of FST decoding is based on \newcite{Koehn2004}, which describes the Pharaoh decoder, now the default decoder in the Moses toolkit.

The biggest challenge for FST decoding is reordering the words in the target language. Assuming an output sentence has $M$ words, there are $M!$ permutations of these words, too many for the decoder to analyze even if it was able to find the correct translation for each word. Instead, FST decoders take a different approach and generate the target sentence word by word, while keeping track of the source words responsible for generating the partial translation so far. This trick effectively swaps the reordering step with the translation step: the algorithm iterates through the source phrases in the order in which their translations would occur in the target language and translates them one by one. The algorithm needs to remember the source words that have already been translated so it can avoid translating them multiple times. Assuming the source sentence has $N$ words, the most compact form in which the decoder can keep track of the processed source words is a $N$ bit mask, where the already translated words are marked with 1 and the remaining words are marked with 0. For example, for the source sentence \textit{So it can happen anywhere}, a 01110 mask implies that the words \textit{it}, \textit{can} and \textit{happen} have already been translated (in any order). This trick reduces the number of reordering states processed by the decoder to $2^N$.

In order to apply target language modeling features, the decoder must have access to the most recent words in each translation hypothesis. For the n-gram back-off language models from \autoref{section:smt:language_models} or the feedforward neural language models discussed in \autoref{chapter:nlms}, the decoder must store the last $n-1$ target words with each state (bit mask). The fact that we do not need to store entire translation hypotheses and that the source coverage bit masks and the $n-1$ target word histories are sufficient leads to the first key optimization applied in FST decoders, known as hypothesis recombination. If two partial hypotheses share the same source coverage bit mask and finish with the same $n-1$ target words, we only need to store the highest scoring one, because for any sequence of future translation rules, this hypothesis will continue to have a higher score. This optimization does not degrade the quality of the decoder.

\begin{figure}
\begin{center}
\includegraphics[scale=0.45]{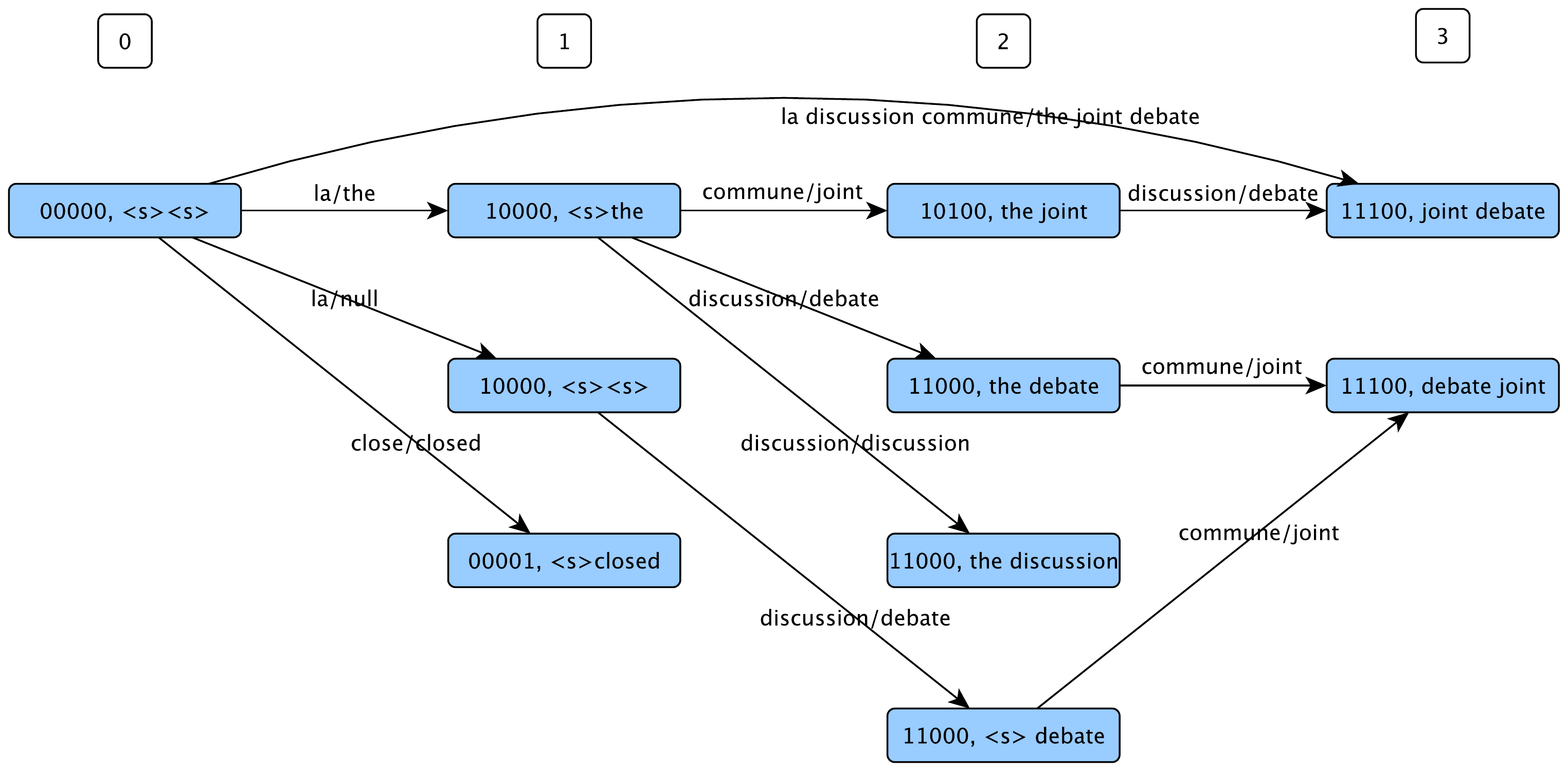}
\caption[Phrase-based decoding]{A subset of the derivations applied by a phrase based decoder with a 3-gram language model to translate the sentence \textit{La discussion commune est close}. The decoder states are stacked in priority queues based on the number of source words translated so far. The transition (10000, $\langle$s$\rangle$ the) $\rightarrow$ (10100, the joint) illustrates a reordering applied by the decoder, where the word \textit{discussion} is left to be translated later. The transitions (00000, $\langle$s$\rangle$ $\langle$s$\rangle$) $\rightarrow$ (11100, joint debate) and (10100, the joint) $\rightarrow$ (11100, joint debate) and, respectively, (11000, the debate) $\rightarrow$ (11100, debate joint) and (11000, $\langle$s$\rangle$ debate) $\rightarrow$ (11100, debate joint) illustrate the hypothesis recombination trick. In the first case, the same translation \textit{the joint debate} is generated by two distinct hypotheses with different scores; in the second case, the hypotheses correspond to different translations: \textit{the debate joint} and \textit{debate joint}. In both cases, the decoder discards the lowest scoring hypothesis.}
\label{fig:fst_decoding}
\end{center}
\end{figure}

The number of decoder states after applying the hypothesis recombination trick is bounded by $O(2^N \times |V_t|^{n-1})$ and the total complexity is $O(2^N \times N \times |V_t|^{n-1})$. Although many target word histories are not valid under the translation model, there are still too many states for the decoder to process. To address this problem, FST decoders limit the size of the window of source words in which the reordering can take place. This by itself is not enough, and another lossy optimization trick known as beam search is employed (\autoref{fig:fst_decoding}). Translation hypotheses, identified by their source bit mask and their $n-1$-gram history, are stacked in priority queues based on the number of covered source words. The decoder iterates over these priority queues in increasing order of covered source words. From each queue, it processes only the top $K$ candidates or those candidates with score above a certain threshold. When choosing which hypotheses to process, the decoder does not rely only on the score given by the scoring model, but it also includes a heuristic score which evaluates how hard it is to translate the remaining part of the source sentence. The purpose of the heuristic is to prevent the decoder from keeping only hypotheses that start with the easy-to-translate parts of the source sentence. The heuristic score is precomputed assuming no reordering is needed for the remaining (uncovered) source words.

The top scoring hypothesis from the priority queue spanning the entire source sentence corresponds to the highest scoring translation. In order to reconstruct the actual translation, for each partial hypothesis we need to store an arc to its parent hypothesis (the hypothesis on which the last derivation was applied before obtaining the current hypothesis). The output sentence is obtained by traversing these arcs starting with the top scoring full hypothesis and by accumulating the target phrases from each derivation in reverse order. If a $n$-best list is needed instead, we can find the $n$ highest scoring paths in this DAG in polynomial time using dynamic programming.

\subsection{Decoding with SCFG Models}\label{subsection:scfg_decoding}

SCFG decoding algorithms are similar to parsing algorithms for CFGs. Their exact formulation depends on the translation model in question, but they share the same general approach. Our exposition is based on hierarchical phrase based models (\autoref{subsection:scfg}).

\begin{figure}
\begin{center}
\includegraphics[scale=0.45]{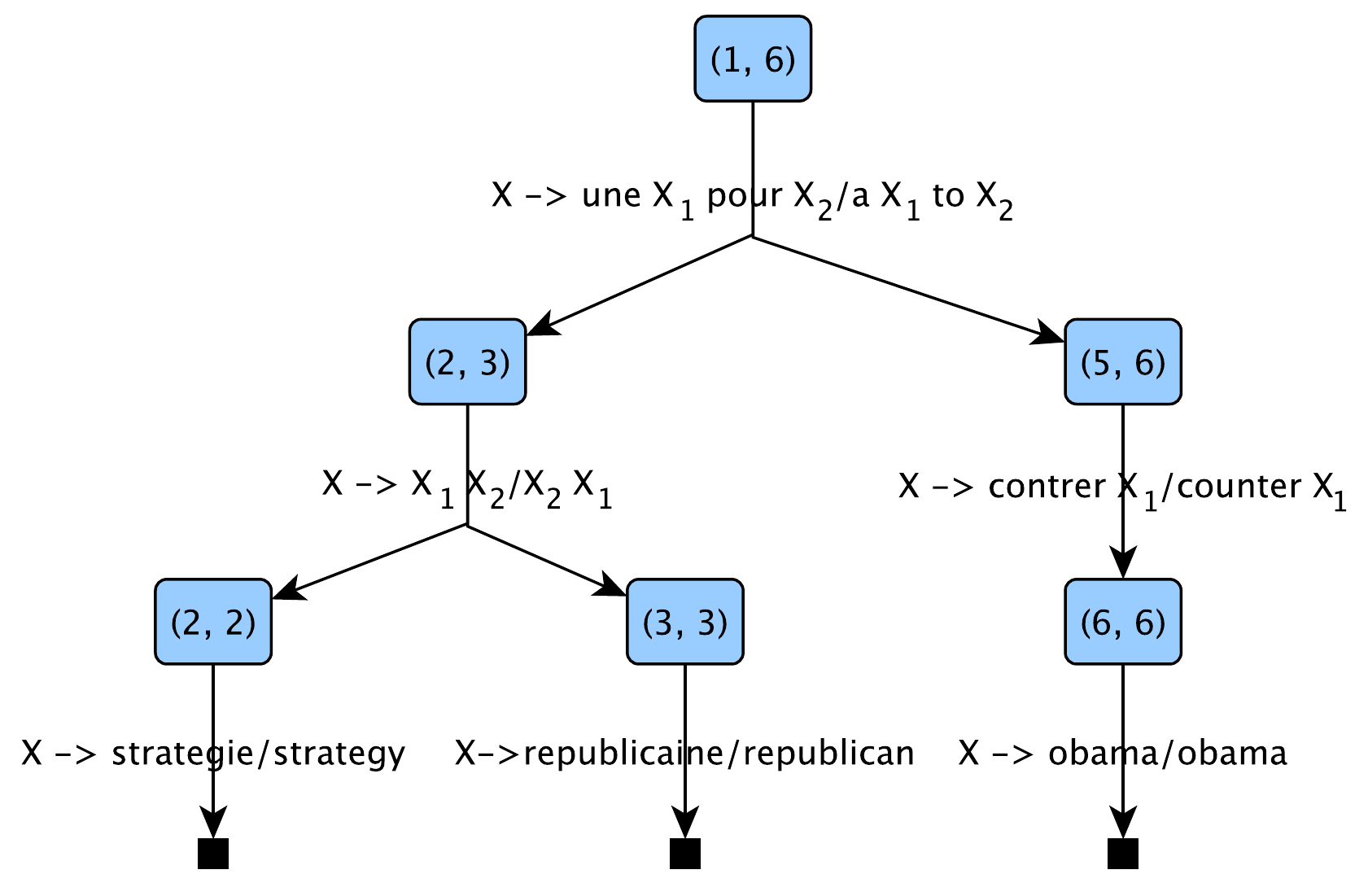}
\caption[Hierarchical phrase-based decoding]{A full derivation obtained with a hierarchical phrase-based decoder for the newspaper headline \textit{A strat\'{e}gie r\'{e}publicaine pour contrer Obama}. Each node is marked with a subspan in the source sentence covered by the decoder state. Reorderings (e.g. \textit{strat\'{e}gie r\'{e}publicaine} $\rightarrow$ \textit{republican strategy}) are delayed until reconstructing the optimal derivation, but these swaps are accounted for in the additional state needed for applying target language models.}
\label{fig:scfg_decoding}
\end{center}
\end{figure}

The decoding algorithm for hierarchical phrase based models \cite{Chiang2007} is a bottom-up dynamic programming algorithm. The algorithm translates contiguous intervals of words from the source sentence in increasing order of their length (span). A state in the decoder is characterized by two indexes $(i, j)$, $1 \leq i \leq j \leq N$, representing the cost of the best partial hypothesis spanning the source words $s_i^j$. To compute the state $(i, j)$, the decoder applies each of the grammar rules matching the source context $s_i^j$, weights them using its scoring model (\autoref{section:smt:scoring_model}) and combines them with any subspans covered by the nonterminals on the right hand side of the rule (\autoref{fig:scfg_decoding}). The derivations licensed by hierarchical phrase-based models contain at most 2 pairs of nonterminals on their right hand side (\autoref{subsection:scfg}). Rules containing 0 or 1 nonterminal pairs can match $s_i^j$ in only one way. For rules containing 2 nonterminal pairs, the algorithm varies the length of the subspan covered by the first nonterminal, which then uniquely identifies the subspan covered by the second nonterminal. For such rules, $O(N)$ pairs of decoder states are analyzed. The overall complexity of SCFG decoding algorithms is cubic in the length of the source sentence, a major improvement over the exponential complexity of FST decoders.

The key challenge with SCFG decoding is incorporating the target language modeling features. In order to apply a n-gram language model, the decoder must store the leftmost and the rightmost $n-1$ target words for any partial hypothesis because new target words can be added at both ends of the hypothesis. After applying the hypothesis recombination trick explained in \autoref{subsection:fst_decoding}, the number of decoder states is $O(N^2 \times |V_t|^{2n-2})$ and the total complexity is $O(N^3 \times |G| \times |V_t|^{2n-2})$. Compared to FST decoders, SCFG decoders only analyze a polynomial number of states in the source sentence, but they store an additional history of $n-1$ target words.

\begin{figure}
\begin{center}
\begin{subfigure}{.33\textwidth}
\begin{tabular}{cc|ccc}
 & & h$_1$ & h$_2$ & h$_3$ \\
 & & 5 & 2 & 1 \\
\hline
 h'$_1$ & 4 & \cellcolor{gray!50}9 & & \\
 h'$_2$ & 3 & & & \\
 h'$_3$ & 1 & & & \\
\end{tabular}
\end{subfigure}
\begin{subfigure}{.33\textwidth}
\begin{tabular}{cc|ccc}
 & & h$_1$ & h$_2$ & h$_3$ \\
 & & 5 & 2 & 1 \\
\hline
 h'$_1$ & 4 & \cellcolor{green!20!blue!35!white}9 & \cellcolor{gray!50}6 & \\
 h'$_2$ & 3 & \cellcolor{gray!50}8 & & \\
 h'$_3$ & 1 & & & \\
\end{tabular}
\end{subfigure}

\begin{subfigure}{.33\textwidth}
\begin{tabular}{cc|ccc}
 & & h$_1$ & h$_2$ & h$_3$ \\
 & & 5 & 2 & 1 \\
\hline
 h'$_1$ & 4 & \cellcolor{green!20!blue!35!white}9 & \cellcolor{gray!50}6 & \\
 h'$_2$ & 3 & \cellcolor{green!20!blue!35!white}8 & \cellcolor{gray!50}5 & \\
 h'$_3$ & 1 & \cellcolor{gray!50}6 & & \\
\end{tabular}
\end{subfigure}
\begin{subfigure}{.33\textwidth}
\begin{tabular}{cc|ccc}
 & & h$_1$ & h$_2$ & h$_3$ \\
 & & 5 & 2 & 1 \\
\hline
 h'$_1$ & 4 & \cellcolor{green!20!blue!35!white}9 & \cellcolor{green!20!blue!35!white}6 & \cellcolor{gray!50}5 \\
 h'$_2$ & 3 & \cellcolor{green!20!blue!35!white}8 & \cellcolor{gray!50}5 & \\
 h'$_3$ & 1 & \cellcolor{gray!50}6 & & \\
\end{tabular}
\end{subfigure}
\caption[Cube pruning]{The first steps of the cube pruning algorithm for two priority queues of hypotheses. Each line corresponds to a hypothesis from the first queue, each column corresponds to a hypothesis from the second queue. Each hypothesis is accompanied by its log-score. The cells in the table correspond to pairs of hypotheses marked by the sum of their individual scores. The priority queue of candidate pairs is shown in gray, while the pairs selected so far are shown in blue. For each selected pair, we insert its two neighboring cells (right and down) in the priority queue. At each step, only the highest scoring pairs are explored.}
\label{fig:cube_pruning}
\end{center}
\end{figure}

In order to make SCFG decoders practical, we apply an optimization trick similar to beam search known as cube pruning. For each pair of source indexes $(i, j)$, we stack the partial hypotheses (identified by their two $n-1$ target word histories) in a priority queue. As with beam search, only the top candidates from each priority queue are analyzed, hypotheses below a certain index or a certain score threshold are discarded. The additional complexity for decoding with hierarchical phrase-based models is that a hypothesis can have up to two parent hypotheses, depending on the number of nonterminal pairs in the top level rule. For rules with two nonterminal pairs, we would like to avoid computing the cartesian product between the priority queues storing the parent hypotheses. Instead, we insert the pair of top candidates from each queue into a new priority queue, identified by its indexes $(1, 1)$. Then, as long as we need to generate new hypotheses, we extract the top pair $(i, j)$ from the new priority queue and insert the pairs $(i+1, j)$ and $(i, j+1)$ avoiding duplicates (\autoref{fig:cube_pruning}). This algorithm guarantees to combine only the top scoring pairs. The number of processed (inserted or extracted) pairs is linear in the size of the output and the operations performed on the additional priority queue have logarithmic time complexity. Overall, the complexity of combining parent hypotheses is reduced from $O(K^2)$ to $O(K \log K)$. This trick can also be generalized to SCFG models where the number of nonterminal pairs in a rule is unlimited.

SCFG decoders use the same algorithm for reconstructing the highest scoring translation (or $n$-best list) from the dynamic programming table as FST decoders.

\section{Evaluation}\label{section:smt:evaluation}

Evaluating the quality of a machine translation system is a hard problem because sentences can often be translated in many ways. It is possible for equivalent translations  not to share any words, while for sentences with many words in common to have completely different meanings. Human translators are able to judge when a system produces good translations, but a more scalable solution was needed to support the massive investments in machine translation over the last decades. As a result, the research community defined and adopted several automatic metrics for evaluating translation quality. Despite their controversy, automatic metrics have several obvious benefits over human evaluation: (i) they facilitate quick iteration by making it easy to see which new features yield improvements, (ii) they are a cost effective way of comparing systems developed by different research groups (assuming the same training and test data is used) and (iii) they eliminate the subjective bias and errors inherent to human evaluation. Quality metrics are also central to feature tuning algorithms like MERT (\autoref{section:smt:scoring_model}), and optimizing for more accurate evaluation metrics can theoretically produce higher quality systems.

Machine translation systems are evaluated on test parallel corpora separated from training and development data in order to prevent overfitting. The decoder translates the source side of the test corpus, and a similarity score is computed between the output translations and the target side of the test corpus (also known as the reference corpus), usually by means of string matching techniques. Evaluation metrics need to be extendable to cases where multiple reference translations are available for each source sentence.

Throughout this thesis, we use BLEU \cite{Papineni2002} to report on translation quality. BLEU is by far the most widely used evaluation metric in the research literature. Other metrics that have seen considerable success include METEOR \cite{Banerjee2005} and TER \cite{Snover2006}. 

BLEU is a precision based metric defined on a $[0, 1]$ scale, where $0$ indicates no overlap between an output translation and its references, while $1$ corresponds to the ideal case where the sentence matches its references exactly. At a high level, BLEU measures how many n-grams from the output translation occur in the reference translations. It defines a modified $n$-gram precision term $p_n$, for every $n \leq 4$, as:
\begin{equation}
p_n = \frac{\sum_{\textbf{t}} \sum_{\textbf{g} \in \text{n-grams(\textbf{t})}} \min(c(\textbf{g}), c_{\text{ref}}(\textbf{g}))}{\sum_{\textbf{t}} \sum_{\textbf{g} \in \text{n-grams(\textbf{t})}} c(\textbf{g})},
\end{equation}
where $c(\textbf{g})$ is the number of times the n-gram $\textbf{g}$ occurs in the output translation $\textbf{t}$ and $c_{\text{ref}}(\textbf{g})$ is the maximum number of times $\textbf{g}$ occurs in any reference translation of $\textbf{t}$. Each term $p_n$ penalizes the translation $\textbf{t}$ if a n-gram $\textbf{g}$ does not occur enough times in $\textbf{t}$. The numerator is capped by $c_{\text{ref}}(\textbf{g})$ in order to prevent falsely increasing the precision by overgenerating words from the reference translations. For robustness, each factor $p_n$ is computed by summing the capped counts and the reference counts over the entire test corpus. The factors $p_n$ are mixed together using the geometric mean.

In order to be a useful metric, BLEU must also address recall. For example, all the factors $p_n$ become 1 if the decoder strictly generates a 4-gram from the reference translations. BLEU addresses the recall problem by penalizing sentences shorter than the reference translations. For each output translation, we consider the reference translation with length closest to that of the output. Let $c$ be the total length of the output translations and $r$ be the total length of the chosen references. BLEU defines a brevity penalty as:
\begin{equation}
BP = \begin{cases}
    1, & \text{if $c > r$},\\
    e^{1-\frac{r}{c}}, & \text{otherwise}.
  \end{cases}
\end{equation}
The brevity term has no effect if the output translations are already longer than the reference translations. Otherwise, the penalty increases exponentially as the output becomes shorter.

Finally, BLEU is defined as the product between the brevity term and the averaged n-gram precision:
\begin{equation}
BLEU = BP \times \exp (\frac{1}{4} \sum_{n \leq 4} \log p_n).
\end{equation}
In order to make gains easier to observe, BLEU is commonly scaled by a factor of $100$. We follow this practice in our work.

\chapter{Online Grammar Extractors}\label{chapter:extractors}

\section{Introduction}

Phrase tables are the default approach for representing translation models in memory. As explained in \autoref{subsection:phrase_tables}, phrase tables load all the phrase pairs extractable from a parallel corpus in memory and organize them as a dictionary, mapping source phrases to lists of target phrases. Phrase tables are very efficient to query because they support constant time access to translation rules, but their main weakness is their huge memory footprint which makes them unsuitable for memory constrained environments. The problem is further aggravated in SCFG-based systems, where the number of extractable rules is exponential in the maximum span of a phrase. In fact, scaling phrase tables for hierarchical phrase-based systems is problematic even without imposing any additional memory constraints.

A naive solution frequently employed by the research community to facilitate decoding with limited resources (e.g. on commodity machines) is to filter the phrase table and remove all translation rules that are not applicable for a given test set. This approach is not satisfactory in our case, because it does not scale to unseen sentences which are to be expected in any practical setting.

In the research literature, a few scalable, compact alternatives to phrase tables have been proposed. \newcite{Zens2007} store phrase tables on disk organized in a trie data structure for efficient read access. \newcite{CallisonBurch2005} and \newcite{Zhang2005} introduce a phrase extraction algorithm based on suffix arrays which extracts translation rules on the fly during decoding. \newcite{Lopez2007} shows how online extractors based on suffix arrays can be extended to extract hierarchical translation rules. 

In our work, we choose to rely on online grammar extractors for retrieving translation rules during decoding. Phrase tables stored on disk and suffix array extractors have comparable lookup times, despite the former having better asymptotic complexity (constant vs. logarithmic), because reading from disk  is slower. However, the suffix array approach yields several practical benefits for our particular setup. First, the amount of disk space available on a mobile device would continue to be an inconvenient limitation if we choose to store phrase tables on disk. Second, assuming we aggressively prune the tables to fit in the available space (and thereby also degrade the model), the initial cost of downloading the models is far greater with this approach. Finally, in order to maintain a manageable size, phrase tables must limit the maximum number of words spanned by a phrase.\footnote{\newcite{Koehn2003} recommend setting the maximum width of a phrase to 4 words.} The memory footprint of online grammar extractors does not depend on this parameter, allowing decoders to use longer phrase pairs, resulting in more accurate translations overall.

In the remainder of this chapter, we discuss an efficient and compact suffix array extractor that works for both standard and hierarchical phrase-based systems. \autoref{section:extractor:contiguous} reviews how suffix arrays are used for contiguous phrase extraction \cite{Lopez2007}. \autoref{section:extractor:discontiguous} introduces a novel algorithm for extracting phrases with gaps. \autoref{section:extractor:implementation} presents details regarding our open source implementation, released as part of the cdec toolkit \cite{Dyer2010}. \autoref{section:extractor:experiments} illustrates the strengths of our approach with experiments. \autoref{section:extractor:summary} concludes with a summary of the ideas discussed in this chapter.

\section{Grammar Extraction for Contiguous Phrases}\label{section:extractor:contiguous}

\begin{figure}
\begin{center}
\begin{tabular}{c|cccccccccc}
$i$ & 0 & 1 & 2 & 3 & 4 & 5 & 6 & 7 & 8 & 9 \\
\hline
$\textbf{w}$ & the & dog & chases & the & cat & many & times & around & the & block \\
$\textbf{a}$ & 7 & 9 & 4 & 2 & 1 & 5 & 8 & 3 & 0 & 6
\end{tabular}
\caption[Suffix array example]{A suffix array constructed from a toy sentence. Note that all suffixes starting with the word \textit{the} are located within a compact interval between positions 6 and 8 (inclusive), illustrating the key property of suffix arrays.}
\label{figure:suffix_array}
\end{center}
\end{figure}

A suffix array \cite{Manber1990} is a memory efficient data structure which can be used to efficiently locate all the occurrences of a pattern, given as part of a query, in some larger string (named \textit{text string} in the string matching literature, e.g. \newcite{Gusfield1997}). A suffix array is the list of suffixes in the text string sorted in lexicographical order. Formally, if $\textbf{a} = a_1^N$ is the suffix array of a string $\textbf{w} = w_1^N$, then $a_i$ stores the starting position of the $i$-th smallest suffix in $\textbf{w}$, i.e. $w_{a_{i-1}}^N < w_{a_i}^N, \forall 1 < i \leq N$. Since each suffix of $\textbf{w}$ is encoded by its starting position in $\textbf{a}$, the overall size of the suffix array is linear in the size of $\textbf{w}$. A crucial property of suffix arrays is that all suffixes starting with a given prefix form a compact interval within the suffix array. Formally, for any $1 \leq i \leq j \leq N$, if $w_{a_i}^N$ and $w_{a_j}^N$ share the same prefix string $\textbf{p}$, then $\forall i \leq k \leq j$, $w_{a_k}^N$ starts with the prefix $\textbf{p}$. An example suffix array constructed from a toy sentence is shown in \autoref{figure:suffix_array}.

Suffix arrays are well suited to solve the central problem of contiguous phrase extraction: efficiently matching phrases against the source side of the parallel corpus. Once all the occurrences of a certain phrase are found, translation rules are extracted from a subsample of phrase matches. The rule extraction algorithm (\autoref{subsection:phrase_tables}) is linear in the size of the phrase pattern and adds little overhead to the phrase matching step.

Before a suffix array can be applied to solve the phrase matching problem, the source side of the parallel corpus is preprocessed by replacing words with numerical ids and concatenating all sentences together into a single array. The suffix array is constructed on top of this new array. In our implementation, we use a memory efficient suffix array construction algorithm proposed by \newcite{Larsson2007} having $O(N \log N)$ time complexity.

\begin{figure}
\begin{center}
\includegraphics[scale=0.35]{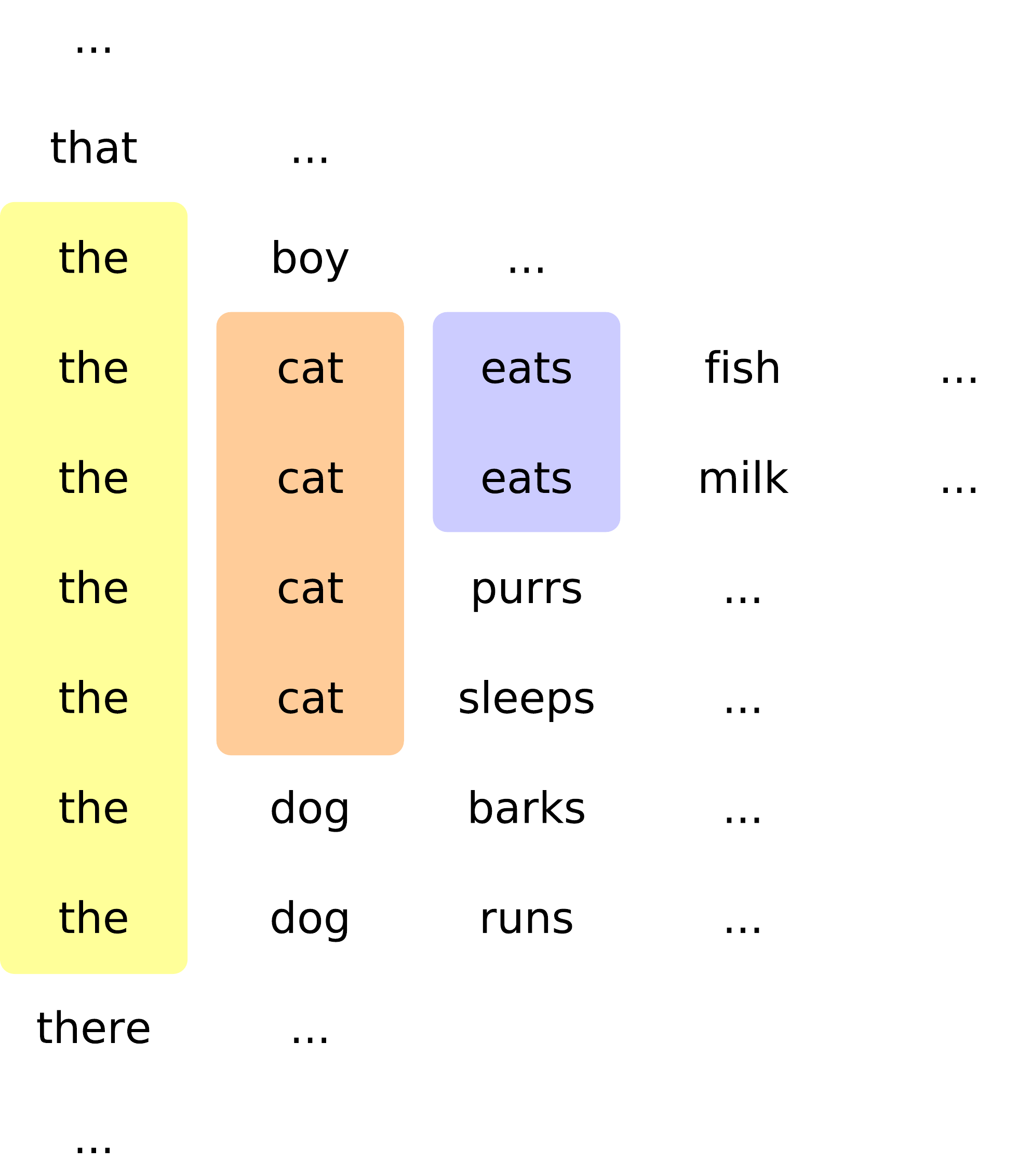}
\end{center}
\caption[Phrase extraction algorithm for contiguous phrases]{A step by step illustration of the contiguous phrase matching algorithm for the phrase \textit{the cat eats}. In the first step, we search for the interval containing all the suffixes starting with the word \textit{the}. In the second step, we continue the search for the phrase \textit{the cat} in this interval, but we only use the second word in each suffix in our comparisons. The third step repeats the operations in the second step, but this time focusing on the word \textit{eats}. Note that in this example we show the suffix array explicitly, but the actual algorithm executes these steps on the compact representation shown in \autoref{figure:suffix_array}.}
\label{figure:contiguous_matching}
\end{figure}

The algorithm for finding the occurrences of a phrase in the parallel corpus uses binary search to locate the interval of suffixes in the suffix array starting with that phrase pattern. Let $w_1, w_2, \ldots, w_K$ be the phrase pattern. Since a suffix array is a sorted list of suffixes, we can binary search for the interval of suffixes starting with $w_1$. This contiguous subset of suffix indices continues to be lexicographically sorted and binary search can be used again to find the subinterval of suffixes starting with $w_1, w_2$. However, all suffixes in this interval are known to start with $w_1$, so it is sufficient to base all comparisons on only the second word in the suffix. The algorithm is repeated until the whole pattern is matched successfully or until the suffix interval becomes empty, implying that the phrase does not exist in the training data. The complexity of the phrase matching algorithm is $O(K \log N)$. The algorithm is illustrated in \autoref{figure:contiguous_matching}.

We note that if $w_1, \ldots, w_{K}$ is a subphrase of a given sentence presented as input to the decoder, then $w_1, \ldots, w_{K-1}$ is also a legitimate subphrase, which the extractor will match as part of a separate query. Matching $w_1, \ldots, w_{K-1}$ executes the first $K-1$ steps of the phrase matching algorithm for $w_1, \ldots, w_K$. Therefore, the complexity of the algorithm can be reduced to $O(\log N)$ per phrase, by caching the suffix array interval found when searching for $w_1, \ldots, w_{K-1}$ and only executing the last step of the algorithm for $w_1, \ldots, w_K$.

\begin{figure}
\begin{center}
\includegraphics[scale=0.6]{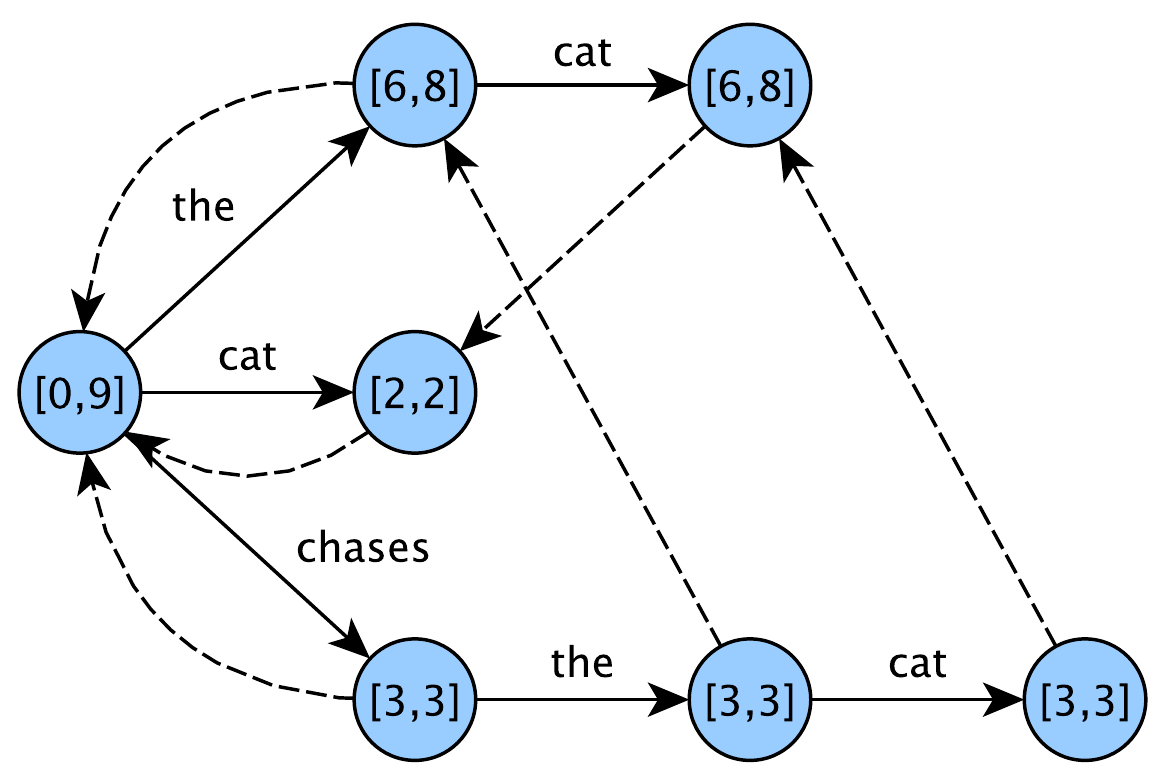}
\end{center}
\caption[Trie cache with suffix links]{The trie cache constructed by algorithm matching contiguous phrases from the input sentence \textit{The cat chases the cat} in the original sentence \textit{The dog chases the cat many times around the block} using the suffix array from \autoref{figure:suffix_array}. Each node corresponds to a matched phrase identified by the path of continuous edges starting at the root of the trie (the leftmost node). For each matched phrase, we label its node with the suffix array interval of suffixes starting with that phrase. Suffix links (shown as dotted lines) unite each phrase of length $n \geq 1$ with its suffix of length $n-1$. We perform a search only if the parent node and the suffix link exist in the trie. The trie is constructed level by level (left to right) and suffix links are updated in $O(1)$ complexity as explained in \newcite{Gusfield1997}.}
\label{figure:trie_cache}
\end{figure}

Let $M$ be the length of a sentence received as input by the decoder. If the decoder explores the complete set of contiguous subphrases of the input sentence, the suffix array is queried $\frac{M (M+1)}{2}$ times. We make two trivial observations to further optimize the extractor by avoiding redundant queries. These optimizations do not lead to major speed-ups for contiguous phrase extraction, but are important for laying the foundations of the extraction algorithm for phrases containing gaps. First, we note that if a certain subphrase of the input sentence does not occur in the training corpus, any phrase spanning this subphrase will not occur in the corpus as well. Second, phrases may occur more than once in a test sentence, but all such repeated occurrences share the same matches in the training corpus. We add a caching layer on top of the suffix array to store the set of phrase matches for each queried phrase. Before applying the pattern matching algorithm for a phrase $w_1, \ldots, w_K$, we verify if the cache does not already contain the result for $w_1, \ldots, w_K$ and check if the search for $w_1, \ldots, w_{K-1}$ and $w_2, \ldots, w_K$ returned any results. The caching layer is implemented as a trie with suffix links and constructed in a breadth first manner so that shorter phrases are processed before longer ones \cite{Lopez2008} (\autoref{figure:trie_cache}). 

\section{Grammar Extraction for Phrases with Gaps}\label{section:extractor:discontiguous}

In \autoref{chapter:smt}, we showed that hierarchical translation systems rely on the synchronous context free grammar formalism which enables them to make use of translation rules containing gaps. In this section, we present an algorithm for extracting synchronous context free rules from a parallel corpus, which requires us to improve the phrase extraction algorithm from \autoref{section:extractor:contiguous} to handle discontiguous phrases. We first published this algorithm in \newcite{Baltescu2014extractor} and it was later adopted as a central piece in \newcite{He2015}'s work to massively scale discontiguous phrase extraction using GPUs. 

Let us make some notations to ease the exposition of the phrase extraction algorithm. Let $a$, $b$ and $c$ be words in the source language, $X$ a nonterminal used to denote the gaps in translation rules and $\alpha$ and $\beta$ source phrases containing zero or more occurrences of $X$. Let $M_{\alpha}$ be the set of matches of the phrase $\alpha$ in the source side of the training corpus, where a phrase match is defined by a sequence of indices marking the positions where the contiguous subphrases of $\alpha$ are found in the training data. Our goal is to find $M_{\alpha}$ for every phrase $\alpha$. \autoref{section:extractor:contiguous} shows how to achieve this if $X$ does not occur in $\alpha$.

Let us now consider the case when $\alpha$ contains at least one nonterminal. If $\alpha = X \beta$ or $\alpha = \beta X$, then $M_{\alpha} = M_{\beta}$, because the phrase matches are defined only in terms of the indices where the contiguous subpatterns match the training data. The words spanned by the leading or trailing nonterminal are not relevant because they do not appear in the translation rule. Since $|\beta| < |\alpha|$, $M_{\beta}$ is already available in the trie cache as a consequence of the breadth first search approach we use to compute the sets $M$.

\begin{figure}
\begin{center}
\includegraphics[scale=0.7]{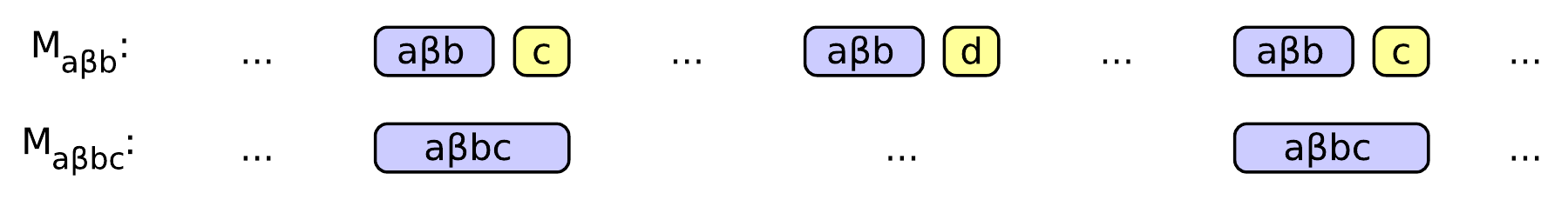}
\caption[Phrase extraction algorithm for discontiguous phrases (case 1)]{The algorithm for constructing $M_{\alpha=a\beta bc}$. For each occurrence of $a\beta b$ in the training data, we check the next symbol and add a new match to $M_{\alpha}$ if the symbol is $c$.}
\label{figure:extractor_case_1}
\end{center}
\end{figure}

\begin{figure}
\begin{center}
\includegraphics[scale=0.7]{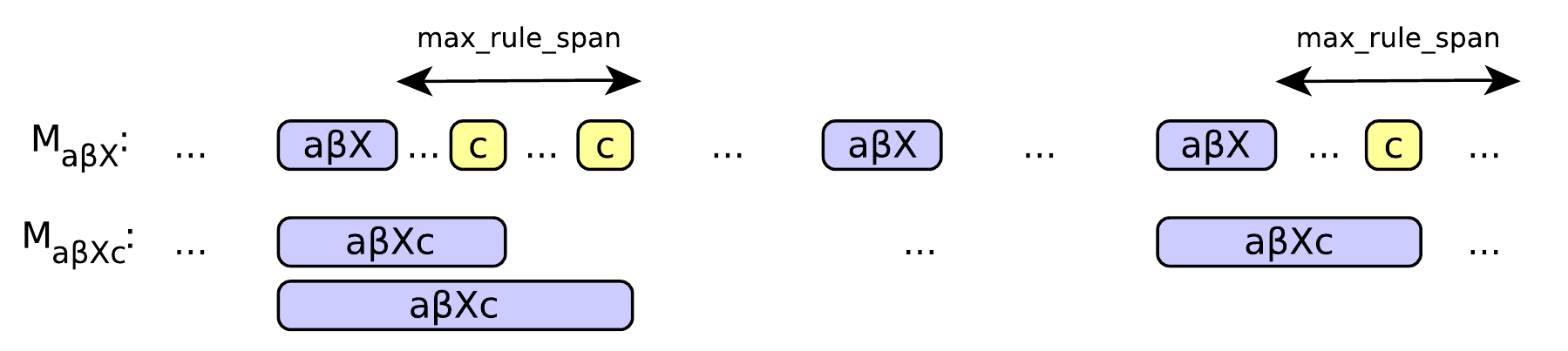}
\caption[Phrase extraction algorithm for discontiguous phrases (case 2)]{The algorithm for constructing $M_{\alpha=a\beta Xc}$. For each occurrence of $a\beta X$ (or $a\beta$) in the training data, we check the next at most \textit{max\_rule\_span} symbols and add one match to $M_{\alpha}$ for every occurrence of $c$. Note that $M_{a\beta X} = M_{a\beta}$ according to our definition of a phrase match.}
\label{figure:extractor_case_2}
\end{center}
\end{figure}

The remaining case is $\alpha = a \beta c$, where both $M_{a \beta}$ and $M_{\beta c}$ have been computed at a previous step. We take into consideration two cases depending on whether the next-to-last symbol of $\alpha$ is a terminal or not (i.e. $\alpha = a \beta b c$ or $\alpha = a \beta X c$, respectively). In the former case, we calculate $M_{\alpha}$ by iterating over all the phrase matches in $M_{a \beta b}$ and selecting those matches that are followed by the word $c$ (\autoref{figure:extractor_case_1}). In the second case, we take note of the experimental results of \newcite{Lopez2008} who shows that translation rules that span more than 15 words have no effect on the overall quality of a translation system. In our implementation, we introduce a parameter \textit{max\_rule\_span} setting the maximum span of a translation rule. For each phrase match in $M_{a \beta X}$, we check if any of the following \textit{max\_rule\_span} words is $c$ (subject to sentence boundaries and taking into account the current span of $a \beta X$) and insert any new phrase matches in $M_{\alpha}$ accordingly (\autoref{figure:extractor_case_2}).

Note that $M_{\alpha}$ can also be computed by considering two cases based on the second symbol in $\alpha$ (i.e. $\alpha = a b \beta c$ or $\alpha = a X \beta c$) and by searching the word $a$ at the beginning of the phrase matches in $M_{b \beta c}$ or $M_{X \beta c}$. In our implementation, we consider both options and apply the one that is likely to lead to a smaller number of comparisons. The complexity of the algorithm for computing $M_{\alpha = a \beta c}$ is $O(\min(|M_{a \beta}|, |M_{\beta c}|))$.

\newcite{Lopez2007} presents a similar grammar extraction algorithm for discontiguous phrases, but the complexity for computing $M_{\alpha}$ is $O(|M_{a \beta}| + |M_{\beta c}|)$. \newcite{Lopez2007} introduces a separate optimization based on double binary search \cite{BaezaYates2004} of time complexity $O(\min(|M_{a \beta}|, |M_{\beta c}|) \log \max(|M_{a \beta}|, |M_{\beta c}|))$, designed to speed up the extraction algorithm when one of the lists is much shorter than the other. Our approach is asymptotically faster than both algorithms. In addition to this, we do not require the lists $M_{\alpha}$ to be sorted, allowing for a much simpler implementation. \newcite{Lopez2007} needs van Emde Boas trees \cite{Cormen2009} and an inverted index to sort these lists efficiently.

The extraction algorithm can be optimized by precomputing an index for the most frequent discontiguous phrases \cite{Lopez2007}. To construct the index, we first identify the most frequent \textit{contiguous} phrases in the training data. We use the LCP array \cite{Manber1990}, an auxiliary data structure constructed in linear time from a suffix array \cite{Kasai2001}, to find all the contiguous phrases in the training data that occur above a certain frequency threshold. We add these phrases to a max-heap together with their frequencies and extract the most frequent $K$ contiguous patterns, where $K$ is a parameter received as input by the grammar extractor. We iterate over the source side of the training data and populate the index with all the discontiguous phrases of the form $u X v$ and $u X v X w$, where $u$, $v$ and $w$ are amongst the most frequent $K$ contiguous phrases in the training data.

\section{Implementation Details}\label{section:extractor:implementation}

Our grammar extractor is designed as a standalone tool which takes as input a word-aligned parallel corpus and a test set and produces as output the set of translation rules applicable to each sentence in the test set. The extractor produces the output in the format expected by the cdec decoder, but the implementation is self-contained and easily extendable to other hierarchical phrase-based translation systems.

Our tool performs grammar extraction in two steps. The preprocessing step takes as input the parallel corpus and the file containing the word alignments and writes to disk binary representations of the data structures needed in the extraction step: a dictionary mapping tokens to numerical ids, the source suffix array, the target data array, the word alignment, the precomputed index of frequent discontiguous phrase matches and a translation table storing count based estimates for the conditional probabilities $p(s | t)$ and $p(t | s)$, for every source word $s$ and target word $t$ collocated in the same sentence pair in the training data. cdec uses both phrase level and word level generative translation models as features in the decoder (see \autoref{section:smt:scoring_model} for details). The translation table is needed to efficiently compute word based features online. The preprocessing step needs to be performed only once when extracting grammars for multiple test corpora. The extraction step takes as input the precomputed data structures and a test corpus and produces a set of grammar files containing the applicable translation rules for each sentence in the test set. The cdec decoder expects that all grammars are made available ahead of time, which is why we process each test corpus as a batch.  We do not take advantage that the corpus is known ahead of time and do not apply pruning techniques as commonly done for phrase tables.

The grammar extractor is written in C++. Our implementation leverages the benefits of a multithreaded environment to speed up grammar extraction. The test corpus is dynamically distributed across the number of available threads (specified by the user via the \textit{-{}-threads} parameter). All the data structures computed in the preprocessing step are immutable during extraction and can be effectively shared across multiple threads at no additional time or memory cost. In contrast, cdec's cython extractor implementing the algorithm proposed by \newcite{Lopez2007} uses the \textit{multiprocessing} library for parallelization. The precomputed data structures are copied across all the processes used for extraction, increasing the memory usage by a factor proportional to the number of processes. As a result, the parallelization feature of the cython extractor is not usable when a limited amount of memory is available.

Our code is released together with a suite of unit tests meant to encourage developers to add their own features to our grammar extractor, without fear that their code changes might have unexpected consequences.

\section{Experiments}\label{section:extractor:experiments}

In this section, we present a set of experiments which illustrate the benefits of our new extraction algorithm. We compare our implementation with the cdec cython extractor which implements the algorithm proposed by \newcite{Lopez2007}.  In order to make the comparison fair and to prove that the speed-ups we obtain are indeed a result of our new algorithm, we also report results for a C++ implementation of the algorithm in \newcite{Lopez2007}.

In our experiments, we used the French-English data from the \textit{europarl} corpus, a set of 2M sentence pairs containing a total of 105M tokens. The training data was tokenized, lowercased and sentence pairs with unusual length ratios were filtered out using the corpus preparation scripts available in cdec.\footnote{We followed the instructions at: \url{http://www.cdec-decoder.org/guide/tutorial.html}.} The corpus was aligned with \textit{fast\_align} \cite{Dyer2013} and the alignments were symmetrized using the \textit{grow-diag-final-and} heuristic (\autoref{section:smt:alignment_models}). We extracted translation rules for the \textit{newstest2012} corpus. The test corpus consists of 3,003 sentences and was tokenized and lowercased using the same scripts. All the data used in these experiments is available on the WMT website.\footnote{The website is accessible at: \url{http://statmt.org/wmt15/}.} In all implementations, if more than 300 matches in the parallel corpus are found for a given input phrase, 300 of these matches are deterministically sampled without replacement for the purposes of phrase extraction.

\begin{table}
\begin{center}
\begin{tabular}{cccc}
\toprule
\textbf{Implementation} & \textbf{Programming} & \textbf{Time} & \textbf{Memory} \\
& \textbf{Language} & \textbf{(minutes)} & \textbf{(GB)} \\
\midrule
\newcite{Lopez2007} & cython & 28.518 & 6.4 \\
\newcite{Lopez2007} & C++ & 2.967 & 6.4 \\
Current work & C++ & 2.903 & 6.3 \\
\bottomrule
\end{tabular}
\caption{Results for the preprocessing step.}
\label{tab:preprocessing}
\end{center}
\end{table}

\begin{table}
\begin{center}
\begin{tabular}{cccc}
\toprule
\textbf{Implementation} & \textbf{Programming} & \textbf{Time} & \textbf{Memory} \\
& \textbf{Language} & \textbf{(minutes)} & \textbf{(GB)} \\
\midrule
\newcite{Lopez2007} & cython & 309.725 & 4.4 \\
\newcite{Lopez2007} & C++ & 381.591 & 6.4 \\
Current work & C++ & 75.496 & 5.7 \\
\bottomrule
\end{tabular}
\caption{Results for the phrase extraction step.}
\label{tab:extraction}
\end{center}
\end{table}

\autoref{tab:preprocessing} shows results for the preprocessing step of the three implementations. We note a 10-fold time reduction when reimplementing \newcite{Lopez2007}'s algorithm in C++. We believe this is a case of inefficient programming when the precomputed index is constructed in the cython code and not a result of using different programming languages. Our new implementation does not significantly outperform the C++ reimplementation of the preprocessing step because we construct the same set of data structures.

The second set of results (\autoref{tab:extraction}) show the running times and memory requirements of the extraction step. Our C++ reimplementation of \newcite{Lopez2007}'s algorithm is slightly less efficient than the original cython extractor, supporting the idea that the two programming languages have comparable performance. We note that our novel extraction algorithm is over 4 times faster than the original approach of \newcite{Lopez2007}.

\begin{table}
\begin{center}
\begin{tabular}{cccc}
\toprule
\textbf{Implementation} & \textbf{Programming} & \textbf{Time} & \textbf{Memory} \\
& \textbf{Language} & \textbf{(minutes)} & \textbf{(GB)} \\
\midrule
\newcite{Lopez2007} & cython & 37.950 & 35.2 \\
\newcite{Lopez2007} & C++ & 51.700 & 10.1 \\
Current work & C++ & 9.627 & 6.1 \\
\bottomrule
\end{tabular}
\caption{Results for parallel extraction using 8 processes/threads.}
\label{tab:multithread_extraction}
\end{center}
\end{table}

\autoref{tab:multithread_extraction} demonstrates the benefits of parallel phrase extraction. We repeated the experiments from \autoref{tab:extraction} using 8 processes in cython and 8 threads in C++. As expected, the running times decrease roughly 8 times. The benefits of sharing the data structures in the parallel extraction step are obvious, our new implementation using 29.1 GB less memory.

\section{Summary}\label{section:extractor:summary}

In this chapter, we investigated compact alternatives for phrase tables. We began with a brief exposition of existing techniques and showed why online grammar extractors are a natural choice for the task of constructing an autonomous translation system that can run on a commodity machine or on a mobile device. We first reviewed how suffix array grammar extractors work in phrase-based translation systems and then introduced a novel extraction algorithm for hierarchical phrases that is 4 times faster than \newcite{Lopez2007}. We provided details on our open source implementation and showed how to maximise parallelism without any negative impact on the memory footprint. Finally, we presented several experiments illustrating the benefits of the approach we proposed in this chapter.

\chapter{Neural Language Models}\label{chapter:nlms}

\section{Introduction}

Most translation systems today use standard back-off n-gram models to model the target language. We showed in \autoref{section:smt:language_models} that traditional n-gram models estimate and store a weight for every n-gram in the training data. The amount of data used to train language models has significant impact on the overall translation quality. Monolingual data is cheap to obtain and it has become a standard to train language models on corpora consisting of billions of words. On these large corpora, even the most compact implementations of n-gram models require many gigabytes of memory, making them unsuitable for memory constrained environments.

In this chapter, we focus on neural language models as a compact alternative to traditional n-gram models. Neural language models \cite{Bengio2003} are a more recent class of language models which use neural networks to learn distributed representations for words. Neural language models project words and contexts into a continuous vector space. The conditional probability $P(w_i | w_{i-n}^{i-1})$ is defined to be proportional to the distance between the continuous representation of the word $w_i$ and the context $w_{i-n}^{i-1}$. Neural language models learn to cluster word vectors according to their syntactic and semantic role. The strength of neural language models lies in their ability to generalize to unseen n-grams, because similar words share the probability of following the same context. Neural language models have been shown to outperform n-gram language models using intrinsic evaluation \cite{Chelba2013,Mikolov2011,Schwenk2007} or when used in addition to traditional models in natural language systems such as speech recognizers \cite{Mikolov2011,Schwenk2007}.

Neural language models are much more computationally intensive than back-off n-gram models. Scaling these models is a hard problem that has received a lot of attention in the research community. For instance, every forward pass through the underlying neural network computes an expensive softmax activation in the output layer of the network with a cost proportional to the product between the size of the vocabulary and the size of the word embeddings. This operation is performed for every n-gram scored by the network and for every example used to update the model during training. Several methods have been proposed to alleviate this problem: some applicable only when training the model \cite{Mnih2012,Bengio2008}, while others can also speed up arbitrary queries \cite{Morin2005,Mnih2009}. Scaling neural language models is a topic covered extensively in this chapter.

In machine translation, it has been shown that neural language models improve translation quality if incorporated as additional features in machine translation decoders \cite{Botha2014,Vaswani2013} or if used for $n$-best list rescoring \cite{Schwenk2010}. One problem is that most of the work done in this line of research uses different techniques for scaling neural language models, leaving the question of finding the optimal neural language model architecture for machine translation unanswered. As part of our goal to design a compact and scalable high quality translation system, we aim to address this problem by conducting a thorough investigation of what optimization techniques work best when integrating neural language models in translation systems. 

Unlike previous research, we focus on how neural language models perform when used as the sole language models in translation systems because we cannot afford to use additional back-off n-gram models in memory constrained environments. We believe this constraint allows us to draw more insightful conclusions about the optimal architecture for neural language models in general. We argue that when neural language models are used in addition to back-off n-gram models, most of the language modeling is actually done by the back-off n-gram models, with the neural models only acting as a differentiating factor when the back-off models cannot provide a decisive probability. This argument is based on the results in \autoref{section:nlms:comparison}, which show that a back-off n-gram model trained without any memory constraints clearly outperforms even our largest neural language models, so it is reasonable to expect that when these models are used together in a translation system, most of the benefits come from the back-off n-gram model. Furthermore, when we trained translation systems with both of these features (e.g. for \newcite{Baltescu2015}, Table 5), we observed that the log-linear scoring model (\autoref{section:smt:scoring_model}) learned on average a weight twice as large for the back-off n-gram model compared to the neural model, implying that the log-linear scoring model sees the unconstrained back-off n-gram model as a more reliable signal than the neural model.

\newcite{Vaswani2013} show that directly integrating neural language models in machine translation decoders leads to better results over using these models only for reranking $n$-best lists. This result matches the intuition that the neural language modeling features help decoders converge on a better set of translations during the exploration of the search space. Based on this result, we focus our investigation on the direct integration of neural language models in machine translation decoders.

We use feedforward neural networks as the underlying architecture for our neural language models because it makes the decoder integration much more straightforward. The decoder needs to store only the $n-1$-gram histories for translation hypotheses (\autoref{section:smt:decoding}). Feedforward neural language models can be integrated in any decoder (that supports back-off n-gram models) without making any changes to the decoder itself. \newcite{Auli2014} show that recurrent neural language models can also be integrated in decoders, but to prevent an exponential blow up in the number of decoder states, they only keep the top scoring hypothesis for any $n-1$-gram history.

The chapter is structured as follows. \autoref{section:nlms:setup} introduces the data we use to perform the experiments which lay the grounds for our investigation. \autoref{section:nlms:model} presents the basic architecture of neural language models which we later seek to improve upon. \autoref{section:nlms:normalization} explores several tricks for reducing the amount of computation in the softmax step. \autoref{section:nlms:noise_contrastive} investigates noise contrastive training, a sampling technique which drastically reduces the complexity of training neural language models, and shows how this method can be used in conjunction with the class-based factorization introduced in \autoref{section:nlms:normalization}. In \autoref{section:nlms:diagonal}, we explore diagonal context matrices as a source for further speed improvements. In \autoref{section:nlms:comparison}, we analyze the performance of neural language models and traditional back-off models on a wide memory spectrum and show that neural language models are superior in memory constrained environments. \autoref{section:nlms:direct_features} shows how neural language models can be extended to include direct n-gram features which allow them to learn weights for n-grams from the conditioning context. \autoref{section:nlms:source_conditioning} presents how the conditioning context of neural language models can be extended with words from the source sentence. \autoref{section:nlms:summary} concludes the chapter with a summary of our learnings.

\section{Experimental Setup}\label{section:nlms:setup}

In our experiments, we use data from the 2014 edition of the Workshop in Machine Translation.\footnote{The data is available at: \url{http://www.statmt.org/wmt14/translation-task.html}.} We train standard phrase-based translation systems for French $\rightarrow$ English, English $\rightarrow$ Czech and English $\rightarrow$ German using the Moses toolkit \cite{Koehn2007}.

\begin{table}
\begin{center}
\begin{tabular}{ccc}
\toprule
\textbf{Language pairs} & \textbf{\# tokens} & \textbf{\# sentences} \\
\midrule
fr$\rightarrow$en & 113M & 2M \\
en$\rightarrow$cs & 36.5M & 733.4k \\
en$\rightarrow$de & 104.9M & 1.95M \\
\bottomrule
\end{tabular}
\end{center}
\caption{Statistics for the parallel corpora.}
\label{table:parallel_data}
\end{table}

\begin{table}
\begin{center}
\begin{tabular}{ccc}
\toprule
\textbf{Language} & \textbf{\# tokens} & \textbf{Vocabulary} \\
\midrule
English (en) & 2.05B & 105.5k \\
Czech (cs) & 566M & 214.9k \\
German (de) & 1.57B & 369k \\
\bottomrule
\end{tabular}
\end{center}
\caption{Statistics for the monolingual corpora.}
\label{table:monolingual_data}
\end{table}

We used the \textit{europarl} and the \textit{news commentary} corpora as parallel data for training the translation systems. The parallel corpora were tokenized, lowercased and sentences longer than 80 words were removed using standard text processing tools.\footnote{We followed the first two steps from \url{http://www.cdec-decoder.org/guide/tutorial.html}.} \autoref{table:parallel_data} contains statistics about the training corpora after the preprocessing step. We tuned the translation systems on the \textit{newstest2013} data using minimum error rate training \cite{Och2003b} and we used the \textit{newstest2014} corpora to report uncased BLEU scores averaged over 3 runs.

The language models were trained on the \textit{europarl}, \textit{news commentary} and the 2007-2013 editions of the \textit{news crawl} corpora. The corpora were tokenized and lowercased using the same text processing scripts. The words which did not occur in the target side of the parallel data were replaced with a special $\langle$unk$\rangle$ token.  We report the sizes of the monolingual corpora and of the vocabularies after preprocessing in \autoref{table:monolingual_data}. 

Throughout this chapter, we report results for 5-gram language models, regardless of whether they are back-off n-gram models or neural language models. To construct the back-off n-gram models, we used the compact trie-based implementation available in KenLM \cite{Heafield2011}. This not only enables us to perform a fair comparison between the two types of models when memory is a key limitation (\autoref{section:nlms:comparison}), but also prevents us from facing difficulties fitting these models in the memory of the machines we use to run experiments. When training neural language models, we set the size of the distributed representations to 500, we use 10 negative samples for noise contrastive estimation (\autoref{section:nlms:noise_contrastive}) and we use diagonal context matrices (\autoref{section:nlms:diagonal}), unless otherwise noted. In cases where we report results on only one language pair, the reader should assume we used French$\rightarrow$English data.

\section{Model Description}\label{section:nlms:model}

\begin{figure}
\begin{center}
\includegraphics[scale=0.28]{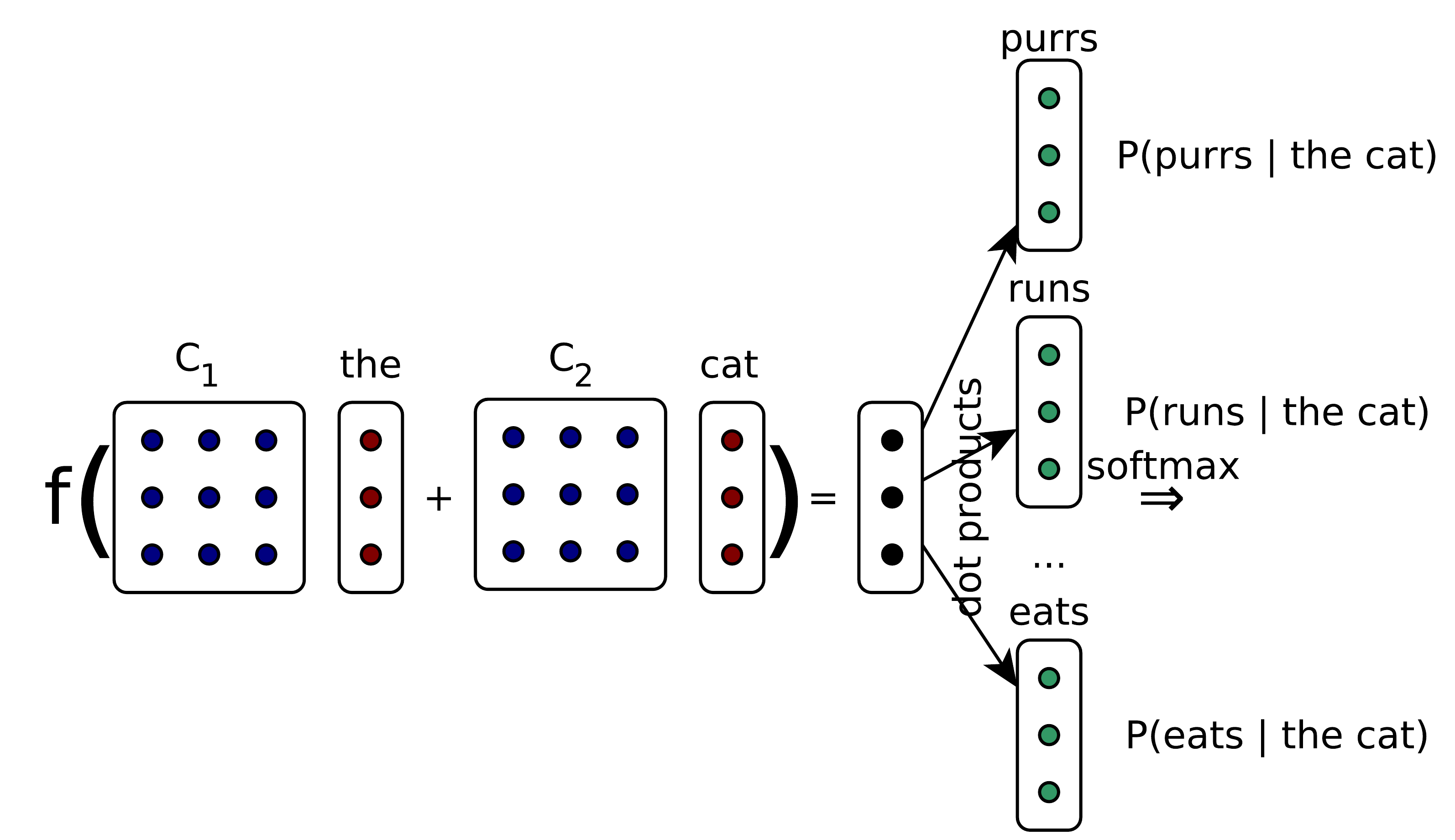}
\caption[Feedforward neural language model architecture]{A 3-gram neural language model is used to predict the word following the context \textit{the cat}.}
\label{figure:architecture}
\end{center}
\end{figure}

As a basis for our investigation, we implement a probabilistic neural language model as defined in \newcite{Bengio2003}.\footnote{We released our implementation as a scalable open source neural language modeling toolkit at: \url{https://github.com/pauldb89/oxlm}.} For every word $w$ in the target vocabulary $V_t$, we learn two distributed representations $\textbf{q}_w$ and $\textbf{r}_w$ in $\mathbb{R}^D$. The vector $\textbf{q}_w$ captures the syntactic and semantic role of the word $w$ when $w$ is part of a conditioning context, while $\textbf{r}_w$ captures its role as a prediction. To simplify our notations, let $h_i$ denote the conditioning context $w_{i-n}^{i-1}$ for a word $w_i$ in a given corpus. To find the conditional probability $P(w_i | h_i)$, our model first computes a context projection vector:
\begin{equation*}
\textbf{p} = f\left( \sum_{j=1}^{n-1} \text{C}_j \textbf{q}_{w_{i-j}} \right),
\end{equation*}
where $\text{C}_j \in \mathbb{R}^{D \times D}$ are context specific transformation matrices and $f$ is a component-wise \textit{rectified linear} activation. The model computes a set of similarity scores measuring how well each word $w \in V_t$ matches the context projection of $h_i$. The similarity score is defined as $\phi(w, h_i) = \textbf{r}_w^\text{T} \textbf{p} + b_w$, where $b_w$ is a bias term initialized with the unigram probability of $w$. A softmax activation is used to transform the similarity scores into probabilities:
\begin{equation*}
P(w_i | h_i) = \frac{\exp (\phi(w_i, h_i))}{\sum_{w \in V} \exp (\phi(w, h_i))}.
\end{equation*}
The model's architecture is illustrated in \autoref{figure:architecture}. 

The complete set of parameters is $(C_j, \textbf{q}_w, \textbf{r}_w, b_w)_{1 \leq j < n,\ w \in V_t}$. The parameters are initialized randomly from a Gaussian distribution with mean 0 and standard deviation 0.1. The model is trained to maximize log-likelihood with $L_2$ regularization.  We shuffle the training data randomly at each iteration and use adaptive gradient descent (AdaGrad) \cite{Duchi2011} to improve convergence. In order to reduce the training time, we split the training data into minibatches of 10 000 samples and evenly distribute the computation of gradient updates within each minibatch over the number of available threads. The model's parameters are updated once at the end of each minibatch.

The architecture presented in this section is not fast enough to scale to real world translation applications. In the next sections, we discuss the key issues with this architecture and investigate several optimizations to address these problems.

\section{Normalization}\label{section:nlms:normalization}

\subsection{Introduction}

Optimizing the computation of the softmax function in the output layer of the network is the most crucial step for making neural language models practical. This operation is performed for every data point presented as input to the network both during training and decoding. For a given input, the time complexity of the softmax step is $O(|V_t| \times D)$. The amount of computation involved is prohibitive given that when decoding with a neural language modeling feature, the model is queried several hundred thousand times for a sentence of average length.

Previous publications on neural language modeling in machine translation have approached this problem in two different ways. \newcite{Vaswani2013} and \newcite{Devlin2014} simply ignore normalization when decoding, albeit \newcite{Devlin2014} alter their training objective to learn self-normalized models, i.e. models where the sum of the values in the output layer is (ideally) close to 1. \newcite{Vaswani2013} use noise contrastive estimation to speed up training, while \newcite{Devlin2014} train their models with standard gradient descent on a GPU.

The second approach is to explicitly normalize the models, but to limit the set of words over which the normalization is performed, either via class-based factorization \cite{Botha2014} or using a shortlist\footnote{Here, the term shortlist should be understood as defined in \newcite{Schwenk2010}.} containing only the most frequent words in the vocabulary and scoring the remaining words with a back-off n-gram model \cite{Schwenk2010}. Hybrid architectures where only infrequent words are grouped into class hierarchies have also been proposed \cite{Le2011}. These normalization techniques can be applied both at training and decoding time.

In this section, we explore both options. The first approach requires little elaboration: the probabilities $P(w | h)$ are replaced by $\phi (w, h)$ when scoring partial hypotheses in the decoder. The second approach is presented in greater detail in the next subsections. \autoref{subsection:nlms:normalization:class_factored_models} introduces class factored models \cite{Botha2014}. \autoref{subsection:nlms:normalization:tree_factored_models} takes this idea further by constructing a multiple level class hierarchy over the vocabulary. \newcite{Schwenk2010}'s approach is not applicable to our particular setup, since it relies on back-off n-gram models for scoring infrequent words. \newcite{Le2011}'s approach does not scale well enough for our goals, since it employs large shortlists and word classes (4000-8000 words), but the core idea is captured in the extension presented in \autoref{subsection:nlms:normalization:tree_factored_models}. We conclude this section with a set of experiments aimed at comparing the effect on speed and translation quality for the optimizations discussed in this section.

\subsection{Class Factored Models}\label{subsection:nlms:normalization:class_factored_models}

\begin{figure}
\begin{center}
\includegraphics[scale=0.6]{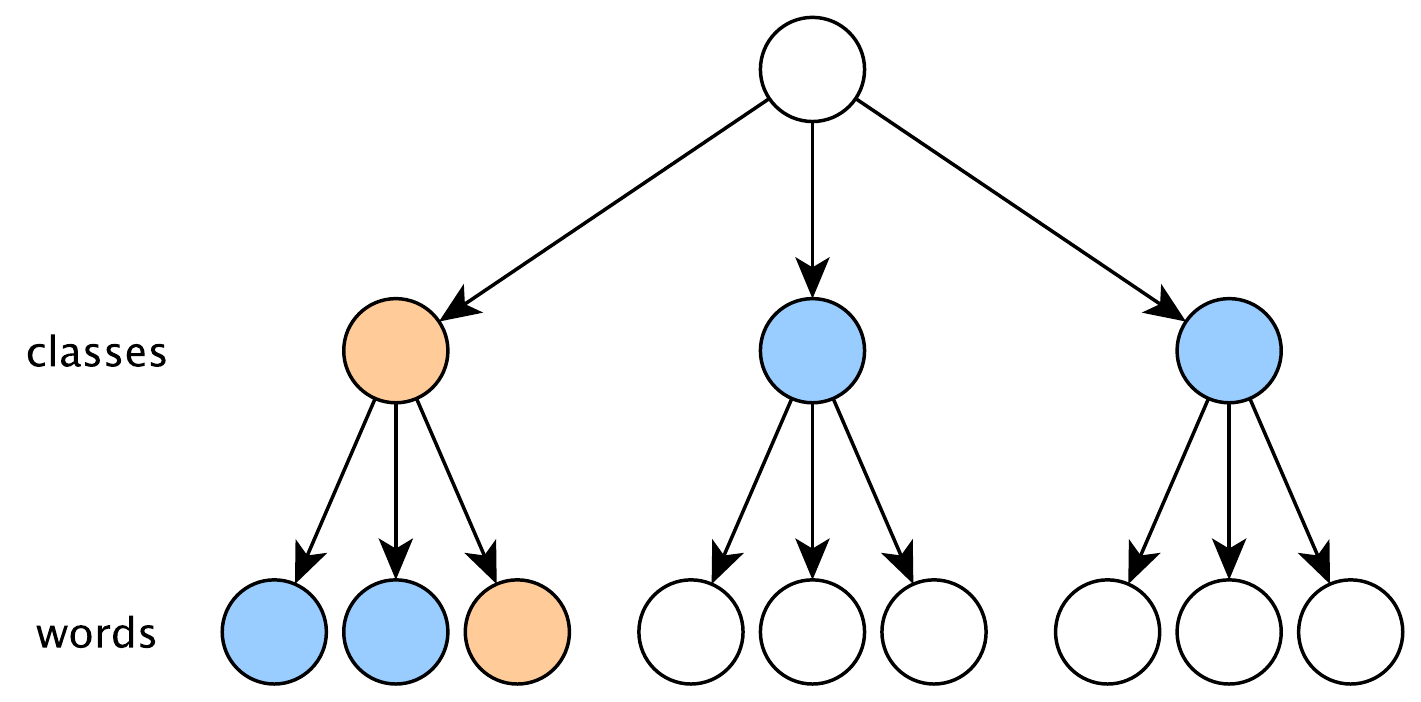}
\caption[Class factored models]{The class factorization in the output layer of the neural network. To compute the probability for the word and the class colored in orange, we need to compute the dot products and normalize over all the colored nodes (both orange and blue). }
\label{figure:class_factored_models}
\end{center}
\end{figure}

The class based factorization trick \cite{Goodman2001} is one option for reducing the excessive amount of computation in the softmax step. We partition the vocabulary into $K$ classes $\{\mathcal{C}_1, \ldots, \mathcal{C}_K\}$ such that $V = \bigcup_{i=1}^K \mathcal{C}_i$ and $\mathcal{C}_i \cap \mathcal{C}_j = \varnothing, \forall 1 \leq i < j \leq K$. We define the conditional probabilities as:
\begin{equation*}
P(w_i | h_i) = P(c_i | h_i) P(w_i | c_i, h_i),
\end{equation*}
where $c_i$ is the class the word $w_i$ belongs to, i.e. $w_i \in \mathcal{C}_{c_i}$. We adjust the model definition to also account for the class probabilities $P(c_i | h_i)$. We associate a distributed representation $\textbf{s}_c$ and a bias term $t_c$ to every class $c$. The class conditional probabilities are computed reusing the projection vector $\textbf{p}$ with a new scoring function $\psi(c, h_i) = \textbf{s}_c^\text{T} \textbf{p} + t_c$. The probabilities are normalized separately:
\begin{align*}
P(c_i | h_i) &= \frac{\exp(\psi(c_i, h_i))}{\sum_{j=1}^K \exp(\psi(c_j, h_i))} \\
P(w_i | c_i, h_i) &= \frac{\exp(\phi(w_i, h_i))}{\sum_{w \in \mathcal{C}_{c_i}} \exp(\phi(w, h_i))}
\end{align*}
\autoref{figure:class_factored_models} illustrates the class decomposition trick. When $K \approx \sqrt{|V_t|}$ and the word classes have roughly equal sizes, the softmax step has a more manageable time complexity of $O(\sqrt{|V_t|} \times D)$ for both training and computing probabilities associated with test n-gram queries.

\subsection{Tree Factored Models}
\label{subsection:nlms:normalization:tree_factored_models}

\begin{figure}
\begin{center}
\includegraphics[scale=0.6]{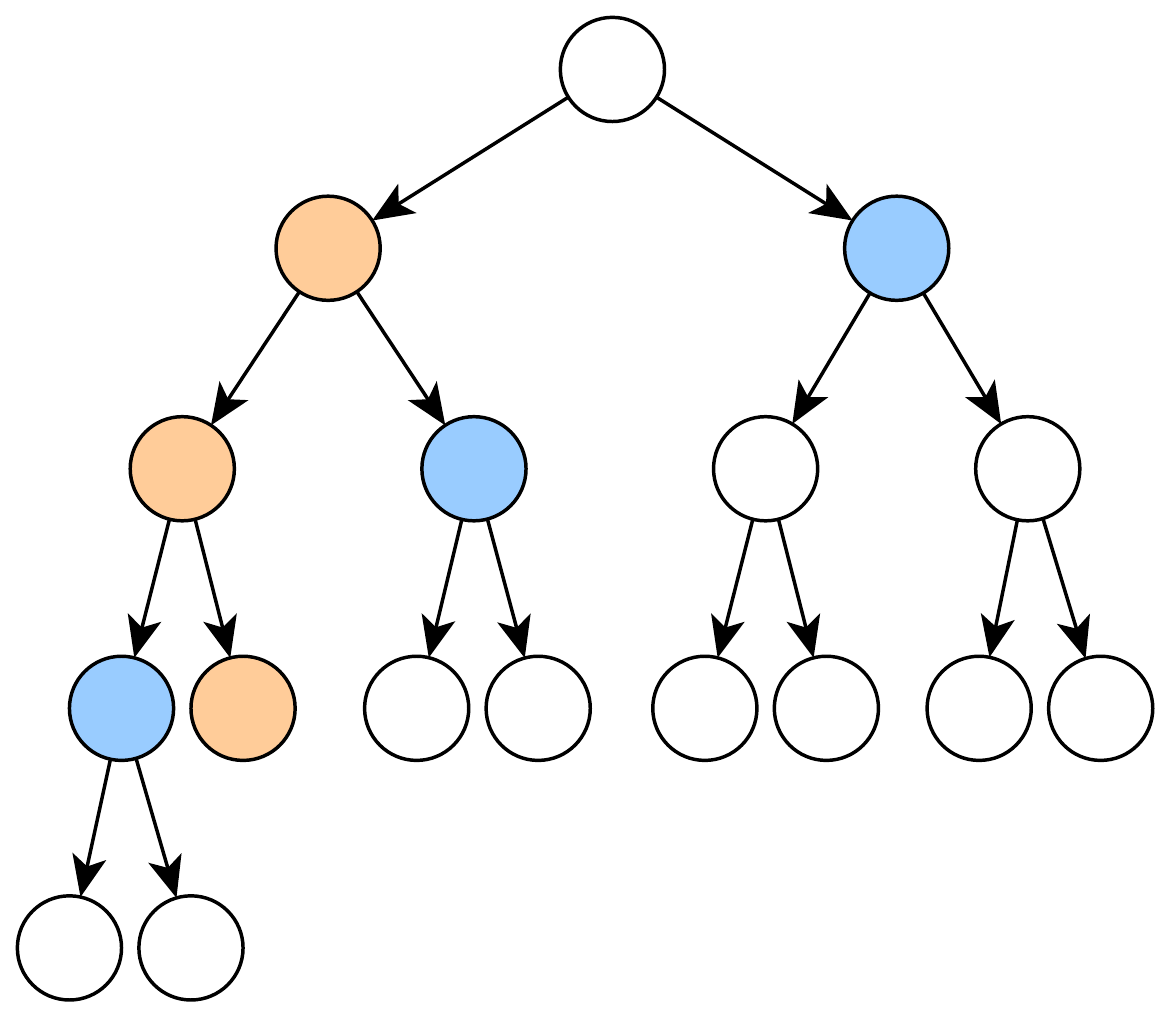}
\caption[Tree factored models]{A multiple level class hierarchy over the vocabulary. To compute $P(w | h)$ for the same word as in \autoref{figure:class_factored_models} (orange path), we need to normalize over all the colored nodes (both orange and blue). }
\label{figure:tree_factored_models}
\end{center}
\end{figure}

One can take the idea presented in the previous subsection one step further and construct a tree over the vocabulary $V_t$. The words in the vocabulary are used to label the leaves of the tree. Let $n_1, \ldots, n_k$ be the nodes on the path descending from the root ($n_1$) to the leaf labelled with $w_i$ ($n_k$). The probability of the word $w_i$ to follow the context $h_i$ is defined as:
\begin{equation*}
P(w_i | h_i) = \prod_{j=2}^k P(n_j | n_1, \ldots, n_{j-1}, h_i).
\end{equation*}
We associate a distributed representation $\textbf{s}_n$ and bias term $t_n$ to each node in the tree. The conditional probabilities are obtained reusing the scoring function $\psi(n_j, h_i)$:
\begin{equation*}
P(n_j | n_1, \ldots, n_{j-1}, h_i) = \frac{\exp(\psi(n_j, h_i))}{\sum_{n \in \mathcal{S}(n_j)} \exp(\psi(n, h_i))},
\end{equation*}
where $\mathcal{S}(n_j)$ is the set containing the siblings of $n_j$ and the node itself. Note that the class decomposition trick described earlier can be seen as a tree factored model with two layers, where the first layer contains the word classes and the second layer contains the words in the vocabulary (see \autoref{figure:class_factored_models} and \autoref{figure:tree_factored_models}).

The optimal time complexity is obtained by using balanced binary trees. The overall complexity of the normalization step becomes $O(\log |V_t| \times D)$ because the length of any path in the tree is bounded by $O(\log |V_t|)$ and because exactly two terms are present in the denominator of every normalization operation. 

Inducing high quality binary trees is a difficult problem which has received some attention in the research literature \cite{Mnih2009,Morin2005}. Results have been somewhat unsatisfactory, with the exception of \newcite{Mnih2009}, who use a top down clustering algorithm for constructing trees. Since \newcite{Mnih2009}'s code has not been open sourced and is difficult to reproduce, in our experiments we use Huffman trees \cite{Huffman1952} instead. Huffman trees do not have any linguistic motivation, but have the property that a minimum number of nodes are inspected during training. In our experiments, the Huffman trees have depths close to $\log{|V_t|}$.

\subsection{Experiments}

\begin{table}[t]
\begin{center}
\begin{tabular}{cccc}
\toprule
\textbf{normalization} & \textbf{fr$\rightarrow$en} & \textbf{en$\rightarrow$cs} & \textbf{en$\rightarrow$de} \\
\midrule
Unnormalized & 30.98 & 18.57 & 18.05 \\
Class Factored  & 31.55 & 18.56 & 18.33 \\
Tree Factored & 30.37 & 17.19 & 17.26 \\
\bottomrule
\end{tabular}
\end{center}
\vspace{-0.3cm}
\caption{BLEU scores for the proposed normalization schemes.}
\label{table:normalization_quality}
\end{table}

In this section, we evaluate three types of models: unnormalized, class factored and tree factored models. These models use diagonal context matrices (\autoref{section:nlms:diagonal}) and are trained using noise contrastive estimation (\autoref{section:nlms:noise_contrastive}), except for tree factored models, where noise contrastive estimation does not improve training time relative to standard gradient descent. In \autoref{table:normalization_quality}, we observe that class factored models perform best in terms of translation quality, with 0.5 BLEU points for fr$\rightarrow$en and 0.3 BLEU points for en$\rightarrow$de over unnormalized models, which are the next best performing models in terms of translation quality. These results also indicate that tree factored models perform poorly compared to the other candidates: -1.2 BLEU points for fr$\rightarrow$en, -1.3 BLEU points for en$\rightarrow$cs and -1 BLEU point for en$\rightarrow$de. We believe this to be a consequence of imposing an artificial hierarchy over the vocabulary.

\begin{table}[t]
\begin{center}
\begin{tabular}{ccc}
\toprule
\textbf{normalization} & \textbf{Clustering} & \textbf{BLEU} \\
\midrule
Class Factored & Brown clustering & 31.55 \\
Class Factored & Frequency binning & 31.07 \\
\bottomrule
\end{tabular}
\end{center}
\vspace{-0.3cm}
\caption{BLEU scores for common clustering strategies on fr$\rightarrow$en data.}
\label{table:clustering}
\end{table}

\begin{table}[t]
\begin{center}
\begin{tabular}{cc}
\toprule
\textbf{Model} & \textbf{Average decoding time} \\
\midrule
KenLM & 1.64 s \\
Unnormalized NLM & 3.31 s \\
Class Factored NLM & 42.22 s \\
Tree Factored NLM & 18.82 s \\
\bottomrule
\end{tabular}
\end{center}
\vspace{-0.3cm}
\caption{Average decoding time per sentence for the proposed normalization schemes.}
\label{table:decoding}
\end{table}

\autoref{table:clustering} compares two popular techniques for obtaining word classes: Brown clustering \cite{Brown1992,Liang2005} and frequency binning \cite{Mikolov2011c}. From these results, we learn that the clustering technique employed to partition the vocabulary can have a huge impact on translation quality and that Brown clustering is clearly superior to frequency binning (0.5 BLEU points for fr$\rightarrow$en).

We report the average time needed to decode a sentence for each of these models in \autoref{table:decoding}. We observe that both class and tree factored models are slow compared to unnormalized models. One option for speeding up factored models is using a GPU to perform the vector-matrix operations. We did not explore this option since GPU implementations are architecture specific and would limit the usability of our open source toolkit by the community. Overall, if quality is more important than speed, we recommend using class factored models; otherwise, we recommend using unnormalized models. 

\section{Noise Contrastive Training}\label{section:nlms:noise_contrastive}

Training neural language models to maximise data likelihood involves several iterations over the entire training corpus and applying the backpropagation algorithm for every training sample. Even with the previous factorization tricks, training neural models is slow. We investigate an alternative approach for training language models based on noise contrastive estimation, a technique which does not require normalized probabilities when computing gradients \cite{Mnih2012}. This method has already been used for training neural language models for machine translation by \newcite{Vaswani2013}.

The idea behind noise contrastive training is to transform a density estimation problem into a classification problem, by learning a classifier to discriminate between samples drawn from the data distribution and samples drawn for a known noise distribution. Following \newcite{Mnih2012}, we set the unigram distribution $P_n(w)$ as the noise distribution and use $k$ times more noise samples than data samples to train our models. The new objective function is:
\begin{equation*}
J(\theta) = \sum_{i=1}^m \log P(C = 1 | \theta, w_i, h_i) + \sum_{i=1}^m \sum_{j=1}^{k} \log P(C = 0 | \theta, n_{ij}, h_i),
\end{equation*}
where $n_{ij}$ are the noise samples drawn from $P_n(w)$. The posterior probability that a word is generated from the data distribution given its context is:
\begin{equation*}
P(C = 1 | \theta, w_i, h_i) = \frac{P(w_i | \theta, h_i)}{P(w_i | \theta, h_i) + k P_n(w_i)}.
\end{equation*}
\newcite{Mnih2012} show that the gradient of $J(\theta)$ converges to the gradient of the log-likelihood objective when $k \rightarrow \infty$. 

When using noise contrastive estimation, additional parameters can be used to capture the normalization terms. \newcite{Mnih2012} fix these parameters to 1 and obtain the same perplexities, thereby circumventing the need for explicit normalization. However, this method does not provide any guarantees that the models are normalized at test time. In fact, the outputs may sum up to arbitrary values, unless the model is explicitly normalized.

Noise contrastive estimation is more efficient than the factorization tricks at training time, but at test time one still has to normalize the model to obtain valid probabilities. We propose combining this approach with the class decomposition trick resulting in a fast algorithm for both training and testing. In the new training algorithm, when we account for the class conditional probabilities $P(c_i | h_i)$, we draw noise samples from the class unigram distribution, and when we account for $P(w_i | c_i, h_i)$, we sample from the unigram distribution of only the words in the class $\mathcal{C}_{c_i}$.

\begin{table}[t]
\begin{center}
\begin{tabular}{cccc}
\toprule
\textbf{Training} & \textbf{Perplexity} & \textbf{BLEU} & \textbf{Duration} \\
\midrule
SGD & 116.596 & 31.75 & 9.1 days \\
NCE & 115.119 & 31.55 & 1.2 days \\
\bottomrule
\end{tabular}
\end{center}
\vspace{-0.3cm}
\caption{A comparison between stochastic gradient descent (SGD) and noise contrastive estimation (NCE) for class factored models on the fr$\rightarrow$en data.}
\label{table:nce}
\end{table}

\begin{table}[t]
\begin{center}
\begin{tabular}{cc}
\toprule
\textbf{Model} & \textbf{Training time} \\
\midrule
Unnormalized NCE & 1.23 days \\
Class Factored NCE & 1.20 days \\
Tree Factored SGD & 1.4 days \\
\bottomrule
\end{tabular}
\end{center}
\vspace{-0.3cm}
\caption{Training times for neural models on fr$\rightarrow$en data.}
\label{table:training}
\end{table}

Class factored models enable us to investigate noise contrastive estimation at a much larger scale than previous results (e.g. the WSJ corpus used by \newcite{Mnih2012} has slightly over 1M tokens), thereby gaining useful insights on how this method truly performs at scale. In our experiments, we use a 2B words corpus and a 100k words vocabulary (\autoref{section:nlms:setup}). \autoref{table:nce} summarizes our findings. We obtain a 0.2 BLEU points improvement using stochastic gradient descent, however, noise contrastive estimation reduces training time by a factor of 7.

\autoref{table:training} reviews the neural models introduced in the previous section and shows the time needed to train each one. We note that noise contrastive training requires roughly the same amount of time for unnormalized and class factored models. Also, we note that this method is at least as fast as maximum likelihood training, even when the latter is applied on tree factored models. Since multi-level tree hierarchies have lower quality, take longer to query and do not yield any substantial benefits at training time compared to unnormalized models, we conclude they represent a suboptimal language modeling architecture for machine translation.

\section{Diagonal Context Matrices}\label{section:nlms:diagonal}

\begin{table}[t]
\begin{center}
\begin{tabular}{cccc}
\toprule
\textbf{Contexts} & \textbf{Perplexity} & \textbf{BLEU} & \textbf{Training time} \\
\midrule
Full & 114.113 & 31.43 & 3.64 days \\
Diagonal & 115.119 & 31.55 & 1.20 days \\
\bottomrule
\end{tabular}
\end{center}
\vspace{-0.3cm}
\caption{A side by side comparison of class factored models with and without diagonal contexts trained with noise contrastive estimation on the fr$\rightarrow$en data.}
\label{table:contexts}
\end{table}

\begin{table}[t]
\begin{center}
\begin{tabular}{ccc}
\toprule
\textbf{Model} & \textbf{Contexts} & \textbf{Average decoding time} \\
\midrule
Factored & Full & 195.74s \\
Factored & Diagonal & 42.22s \\
Unnormalized & Full & 167.45s \\
Unnormalized & Diagonal & 3.31s \\
\bottomrule
\end{tabular}
\end{center}
\vspace{-0.3cm}
\caption{Decoding time speed-ups when using diagonal context matrices.}
\label{table:contexts_decoding}
\end{table}

In this section, we investigate diagonal context matrices as a source for reducing the computational cost of calculating the projection vector \textbf{p}. In the standard definition of a neural language model, this cost is dominated by the softmax step, but as soon as tricks like noise contrastive estimation or hierarchical factorizations are used, this operation becomes the main bottleneck for training and querying the model. In a diagonal matrix, all the elements located outside the main diagonal are zero. As a result, the matrix vector products $\text{C}_j \textbf{q}_{w_{i-j}}$ for computing the projection vector $\textbf{p}$ become scalar products and the time complexity is reduced from $O(D^2)$ to $O(D)$. A similar time reduction is achieved in the backpropagation algorithm, as only $O(D)$ context parameters need to be updated for every training instance.

\newcite{Devlin2014} also identified the need for finding a scalable solution for computing the projection vector. Their approach is to cache the product between every word embedding and every context matrix and to look up these terms in a cache as needed. \newcite{Devlin2014}'s approach works well when decoding, but it is not applicable during training and it requires additional memory. 

\autoref{table:contexts} compares the training time for diagonal and full context matrices for class factored models.
Both models have similar BLEU scores, but the training time is reduced by a factor of 3 when diagonal context matrices
are used. \autoref{table:contexts_decoding} shows the average decoding time per sentence with diagonal and full
context matrices for class factored models and unnormalized models. We observe a time reduction by a factor of 4 for
class factored models and a time reduction by a factor of 50 when the normalization step is skipped altogether.

\section{Quality vs. Memory Trade-Off}\label{section:nlms:comparison}

\begin{table}[t]
\begin{center}
\begin{tabular}{ccc}
\toprule
\textbf{Acceptance Ratio} & \textbf{Memory} & \textbf{BLEU} \\
\midrule
0.01 & 214 MB & 29.78 \\
0.02 & 404 MB & 30.25 \\
0.05 & 922 MB & 31.00 \\
0.1 & 1.7 GB & 31.80 \\
0.2 & 3.1 GB & 32.38 \\
0.5 & 6.6 GB & 32.89 \\
1 & 12 GB & 33.01 \\
\bottomrule
\end{tabular}
\end{center}
\vspace{-0.3cm}
\caption{BLEU scores and memory requirements for back-off n-gram models constructed from different percentages of the training corpus.}
\label{table:acceptance_ratio}
\end{table}

In this section, we compare neural language models and back-off n-gram models on a wide memory spectrum and track how the end to end quality of a translation system changes as these models grow in size. We perform this analysis with two goals in mind. First, we seek to verify the hypothesis that neural language models are indeed better suited for memory constrained environments. Second, we aim to understand how neural language models perform relative to back-off n-gram models when no memory restrictions are enforced, and if increasing the size of the word embeddings beyond what is typically reported in the research literature is sufficient to correct any BLEU score discrepancies.

In this analysis, we used a compact trie-based implementation with floating point quantization for constructing memory efficient back-off n-gram models \cite{Heafield2011}. A 5-gram model trained on the English monolingual data introduced in \autoref{section:nlms:setup} requires 12 GB of memory. We randomly sample sentences with an acceptance ratio ranging between 0.01 and 1 to construct smaller models. The BLEU scores obtained using these models are reported in \autoref{table:acceptance_ratio}.

\begin{table}[t]
\begin{center}
\begin{tabular}{ccc}
\toprule
\textbf{Embeddings Size} & \textbf{Memory} & \textbf{BLEU} \\
\midrule
100 & 86 MB & 29.87 \\
200 & 167 MB & 30.64 \\
500 & 408 MB & 31.55 \\
1000 & 816 MB & 31.97 \\
2000 & 1.6 GB & 32.29 \\
5000 & 4 GB & 32.37 \\
\bottomrule
\end{tabular}
\end{center}
\vspace{-0.3cm}
\caption{BLEU scores and memory requirements for class factored models with different word embeddings sizes.}
\label{table:embedding_size}
\end{table}

We use class factored models with diagonal context matrices as the representative architecture for neural language models. The normalization techniques discussed in \autoref{section:nlms:normalization} and noise contrastive estimation do not affect the memory footprint of these models. Using full context matrices would increase the size of the models, but by not more than 3\%. We do not use any additional compression techniques (e.g. floating point quantization), although this is technically possible. To observe how these models perform for various memory thresholds, we experiment with setting the size of the word embeddings between 100 and 5000. The results are displayed in \autoref{table:embedding_size}.

\begin{figure}
\begin{center}
\includegraphics[scale=0.45]{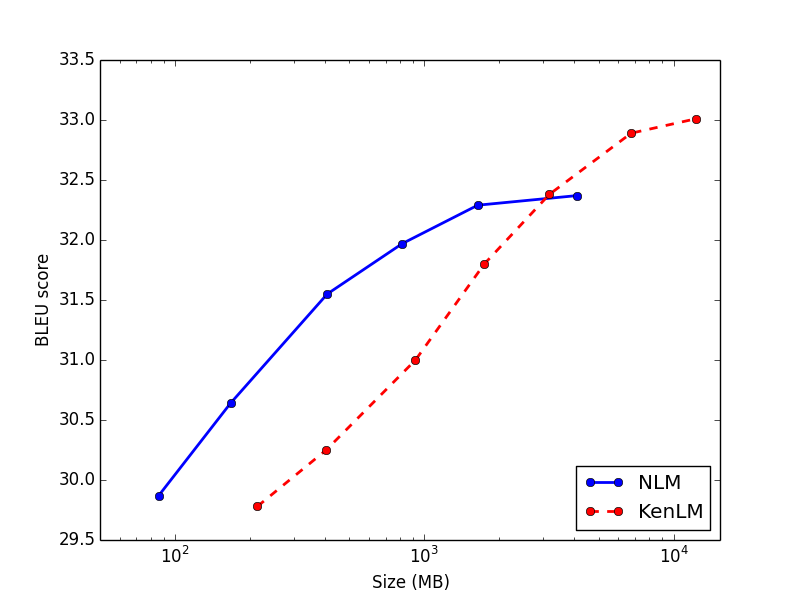}
\caption{BLEU score vs. model size (log scale) for n-gram and neural models.}
\label{fig:memory}
\end{center}
\end{figure}

We plot the quality vs. memory trade-off for back-off n-gram models and class factored models in \autoref{fig:memory}. We observe that neural models perform significantly better (over 1 BLEU point) when less than 1 GB of memory is used, which is roughly the amount of memory available on mobile phones and commodity machines. This result confirms the potential of neural language models for applications designed to run in memory constrained environments. At the other end of the scale, we see that back-off models outperform even the largest neural language models by a decent margin (almost 1 BLEU point) and that we can expect only modest gains if we increase the size of the word embeddings further. This result suggests that increasing the size of the word embeddings is not sufficient for correcting the quality gap between neural language models and back-off n-gram models.

\section{Direct N-gram Features}\label{section:nlms:direct_features}

Neural language models learn embeddings for single words which they combine using context matrices and an activation function. Direct features extend neural language models by allowing them to learn weights for n-grams. Direct features play a similar role to frequency counts in back-off n-gram models: larger predictive weights correspond to more likely n-grams. These weights are used to amend the similarity scores produced by neural language models in cases where the occurrence of several words together in the conditioning context provides more accurate signal.  From the perspective of the underlying neural network, predictive weights are edges connecting the input and the output layers of the network directly.

Direct features for unigrams were originally introduced in neural language models by \newcite{Bengio2003}. \newcite{Mikolov2011} extend these features to n-grams and show they are useful for reducing perplexity and improving word error rate in speech recognizers. Direct n-gram features are reminiscent of maximum entropy language models \cite{Berger1996} and are sometimes called maximum entropy features \cite{Mikolov2011}. 

To our knowledge, prior to our work, neural language models with direct n-gram features have not been explored in machine translation. In this section, our primary goal is to understand the effect of direct n-gram features on translation quality. A common belief in the NLP community is that back-off n-gram models are better at estimating the probabilities of frequent n-grams compared to neural language models. We would like to see what gains are obtained when our models learn the weights of the most frequent n-grams in the training data.

Formally, we define a set of binary feature functions $\textbf{f} = f_1^M$, where $M$ is the number of n-grams we choose to model. We assign each function a weight from a real valued vector $\textbf{u} \in \mathbb{R}^M$. To account for word classes, we also define a set of binary feature functions $\textbf{g}_c$ and a vector of weights $\textbf{v}_c$ for each word cluster $c$. An n-gram $(w, h)$ has a corresponding feature function $g_c(w, h)$ iff $w \in \mathcal{C}_c$. The scoring functions introduced in \autoref{section:nlms:model} are updated to include an additional term:
\begin{align*}
\psi(c, h_i) &= \textbf{s}^T_{c} \textbf{p} + t_{c} + \textbf{u}^T \textbf{f}(c, h_i) \\
\phi(w, h_i) &= \textbf{r}^T_{w} \textbf{p} + b_{w} + \textbf{v}_{c}^T \textbf{g}_{c}(w, h_i)
\end{align*}
In all other respects, our model definition remains unchanged. The weight vectors $\textbf{u}$ and $\textbf{v}_c$ are learned together with the other parameters using noise contrastive estimation.

Fundamentally, neural language models with direct n-gram features require at least as much memory as n-gram language models because they learn and store individual weights for every n-gram. This is why we choose to focus only on the most frequent n-grams in the training data. In our implementation, we need two indexes: one for mapping n-grams to their weights, the other for mapping n-gram contexts $h_i$ to the lists of words and classes that can follow $h_i$. We needed the latter index in order to normalize our models both when computing the gradient at training time and when querying the model at decoding time. We used feature hashing \cite{Weinberger2009,Ganchev2008} to compress these indexes. Research literature shows that as long as the number of collisions resulting from feature hashing is low, the overall quality of the models remains unaffected. We experienced the same outcome when hand tuning the models looking for a good memory vs. quality trade-off.

\begin{table}[t]
\begin{center}
\begin{tabular}{cccc}
\toprule
\textbf{Number of n-grams} & \textbf{Minimum frequency} & \textbf{Memory} & \textbf{BLEU} \\
\midrule
0 & N/A & 408 MB & 31.55 \\
1M & 497 & 511 MB & 31.69 \\
3M & 169 & 621 MB & 32.10 \\
10M & 51 & 0.98 GB & 32.24 \\
30M & 17 & 2.07 GB & 32.38 \\
100M & 5 & 6.25 GB & 32.55 \\
\bottomrule
\end{tabular}
\end{center}
\vspace{-0.3cm}
\caption{BLEU scores and memory requirements for class factored neural language models with n-gram features.}
\label{table:direct_ngrams}
\end{table}

In our experiments, we take the most frequent $M$ n-grams in the corpus, where $M$ varies from 1 million to 100 millions (\autoref{table:direct_ngrams}). For each model, we report the minimum frequency of the n-grams included in the model, together with the memory footprint and the BLEU score. These models are built on top the 500-dimensional class factored model with diagonal contexts which we have used as a baseline for neural language models in this chapter. Overall, we observe a solid improvement in translation quality of 1 BLEU point. If we limit the memory footprint to at most 1 GB,  we get over half a BLEU point improvement. These gains are comparable to those we obtain when increasing the size of the word embeddings. On the other hand, for our largest model, the increase in training and decoding time is roughly 2x, whereas a model with 5000-dimensional embeddings takes 10x more time to train and query compared to the baseline. 

\section{Source Sentence Conditioning}\label{section:nlms:source_conditioning}

In this section, we follow \newcite{Devlin2014} and investigate extending the conditioning context of neural language models with a window of words from the source sentence. This idea is similar to the recent trend in neural translation models \cite{Kalchbrenner2013,Sutskever2014,Cho2014}, however, our models do not capture the full translation process, because we rely on a separate alignment model (\autoref{section:smt:alignment_models}) to locate the relevant source words.

Formally, a source conditioned language model computes the probability of a target sentence \textbf{t} given a source sentence \textbf{s} and an alignment \textbf{a} as follows:
\begin{equation*}
P(\textbf{t} | \textbf{s}, \textbf{a}) = \prod_i P(t_i | t_1^{i-1}, \textbf{s}, \textbf{a}).
\end{equation*}
In our models, we continue to use the traditional $n$-th order Markov assumption on the target side. Also, instead of modeling the entire source sentence, we focus only on a window of most relevant $2m+1$ source words for each target word $t_i$. The window of source words is selected using the alignment links $\textbf{a}$ and is centered around $\bar{a}_i$. Specifically:
\begin{itemize}
\item If $t_i$ has a single alignment link $a_i$, then $\bar{a}_i = a_i$.
\item If $t_i$ has multiple alignment links, $\bar{a}_i$ is chosen to be the middle alignment link. In case of an even number of alignment links, we choose the middle link to the left.
\item If $t_i$ is unaligned, we set $\bar{a}_i = \bar{a}_j$ where $t_j$ is the closest aligned target word to $t_i$. In case of ties, preference is given to the target word on the right.
\end{itemize}
Using these notations, the conditional probabilities become:
\begin{equation*}
P(t_i | t_1^{i-1}, \textbf{s}, \textbf{a}) = P(t_i | t_{i-n}^{i-1}, s_{\bar{a}_i - m}^{\bar{a}_i + m}).
\end{equation*}

\begin{figure}
\begin{center}
\includegraphics[scale=0.5]{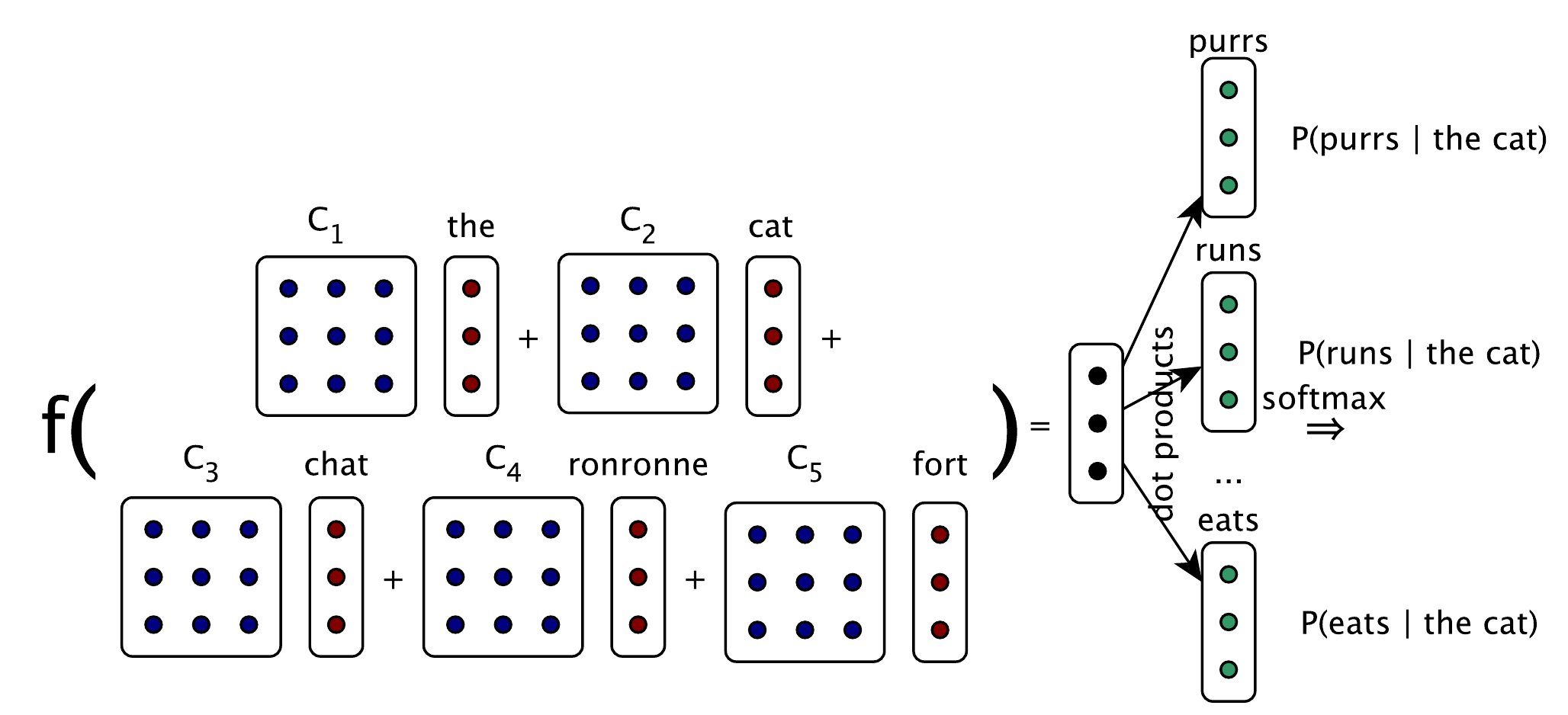}
\caption[Feedforward source conditioned neural language model architecture]{A source conditioned neural language model with $n=3$ and $m=1$ predicting the word following the target context \textit{the cat} using the source window \textit{chat ronronne fort}.}
\label{figure:architecture_source}
\end{center}
\end{figure}

We extend the model presented in \autoref{section:nlms:model} by defining a vector representation $\textbf{q}_w \in \mathbb{R}^D$ for every source word $w \in V_{s}$. These representations capture the syntactic and semantic roles of source words and, together with the embeddings in the target conditioning context, they are used to predict the next target word. We also define $2m+1$ additional (diagonal) context matrices $C_j$. The description from \autoref{section:nlms:model} remains unchanged if we extend the histories $h_i$ to be the concatenation of $t_{i-n}^{i-1}$ and $s_{\bar{a}_i - m}^{\bar{a}_i + m}$. The new model architecture is illustrated in \autoref{figure:architecture_source}.

At decoding time, we need access to the alignment links that served as the basis for extracting each translation rule from the parallel corpus. Fortunately, alignment links are included by default in the phrase tables constructed with the Moses toolkit. cdec's online grammar extractor presented in \autoref{chapter:extractors} also extracts alignment links. The decoder feature applying the source conditioned neural language model uses these links and the location of the source phrase in the source sentence to compute the window of source words used to score each word in the target phrase.

In our experiments, we use the bilingual data from \autoref{section:nlms:setup} to train the language models. We replace the words occurring only once in each side of the corpus with a special $\langle$UNK$\rangle$ token to learn an embedding for unseen words. We use 500 dimensional word embeddings for both source and target words and we train the models using noise contrastive estimation. We used the class based factorization trick to speed up the normalization step. In our experiments, the decoder uses both a monolingual and a source conditioned neural language model as features.

\begin{table}[t]
\begin{center}
\begin{tabular}{cccc}
\toprule
\textbf{Model} & \textbf{fr$\rightarrow$en} & \textbf{en$\rightarrow$cs} & \textbf{en$\rightarrow$de} \\
\midrule
target & 31.55 & 18.56 & 18.33 \\
source + target & 31.81 & 18.71 & 18.64 \\
\bottomrule
\end{tabular}
\end{center}
\vspace{-0.3cm}
\caption{BLEU scores for source conditioned neural language models.}
\label{table:source_conditioned}
\end{table}

The additional source conditioned neural language model results in 0.2-0.3 gains in BLEU score across all language pairs (\autoref{table:source_conditioned}). While these gains may appear modest, we note that the size of the parallel corpora used to train these models is 15-18 times smaller than the monolingual corpora used to train the target language models. On the positive side, these models require about as much memory as the monolingual models despite learning embeddings in two languages because the vocabularies are smaller.

\section{Summary}\label{section:nlms:summary}

In this chapter, we introduced neural language models as a compact alternative to traditional back-off n-gram models. Neural language models are notoriously difficult to scale. We addressed the key structural inefficiencies in these models and analyzed the effect of the proposed optimizations on end to end translation quality. First, we focused on speeding up the softmax step in the output layer of the neural network. We investigated factoring the output layer using word classes and multiple level trees as well as simply ignoring the normalization altogether. We found out that class factored models perform best in terms of quality, but unnormalized models are considerably faster. Tree factored models did not perform nearly as well as the other candidates and did not provide significant speed ups. We also found out that the clustering algorithm used for defining the word classes can have a significant effect on quality and that Brown clustering \cite{Brown1992} is a good choice. We showed that noise contrastive estimation is a fast and effective training algorithm for neural language models that works well for large datasets. We showed how noise contrastive estimation can be extended to class factored models. We also showed that diagonal context matrices can significantly speed up the computation of the hidden layer without considerable impact on quality. Putting this information together, we established a strong baseline for neural language models, which we compared with a compact implementation of back-off n-gram models \cite{Heafield2011}. We observed that class factored models outperform back-off n-gram models by 1 BLEU point when the amount of available memory is limited to 1GB. However, when this restriction is lifted, back-off n-gram models outperform neural language models by a significant margin. Furthermore, increasing the size of the word embeddings is not sufficient to close the gap between these types of models. Finally, we explored direct n-gram features and source sentence conditioning as extensions to neural language models. Both techniques showed reasonable gains in BLEU score.

\chapter{Conclusions}\label{chapter:conclusions}

\section{Building a Compact Translation System}

Our primary goal in this thesis is to produce a scalable, high quality machine translation system that can operate with the limited resources available on mobile devices. The amount of memory available on such devices is commonly limited to 1GB. This is the main bottleneck that prevents standard machine translation systems from being employed on mobile devices. In \autoref{chapter:smt}, we reviewed the components that comprise a standard translation system and identified phrase tables and traditional back-off n-gram language models as the components responsible for rendering these systems impractical in memory constrained environments. \autoref{chapter:extractors} proposed online suffix array grammar extractors as a compact, scalable alternative to phrase tables and introduced a new algorithm that makes phrase extraction considerably faster for hierarchical phrase based systems. \autoref{chapter:nlms} investigated neural language models as a memory efficient alternative to traditional back-off language models. We addressed the key scalability concerns regarding neural language models and explored a few extensions that lead to translation quality improvements. In this section, we use the suffix array grammar extractors and the neural language models introduced in the earlier chapters in a standard translation system and show that our system produces high quality translations with limited memory.

We train hierarchical phrase-based systems with the cdec toolkit \cite{Dyer2010} for three language pairs: fr$\rightarrow$en, en$\rightarrow$cs and en$\rightarrow$de and limit the total memory used by the systems to 1GB. The grammars are extracted using the new, faster algorithm for discontiguous phrases introduced in \autoref{section:extractor:discontiguous}. Despite being more compact than phrase tables, the memory footprint of suffix array extractors is linear in the size of the training data, which limits the amount of parallel data we can use for extracting translation rules. In our experiments, we use the \textit{news commentary} corpora from WMT 2014\footnote{The parallel and monolingual data used in these experiments is available at \url{http://statmt.org/wmt14/}.} for grammar extraction. For language modeling, we replace the default implementation using back-off n-gram models \cite{Heafield2011} available in cdec with the class factored neural language models introduced in \autoref{chapter:nlms}. The output layer of the neural models is factored using Brown clustering \cite{Liang2005}. We use diagonal context matrices and project the word embeddings and the hidden layer into a 500-dimensional vector space. The models are trained using noise contrastive estimation on the \textit{europarl}, \textit{news commentary} and \textit{news crawl} corpora.

\begin{table}
\begin{center}
\begin{tabular}{cccc}
\toprule
\textbf{Language pairs} & \textbf{fr$\rightarrow$en} & \textbf{en$\rightarrow$cs} & \textbf{en$\rightarrow$de} \\
\midrule
SCFG memory & 460 MB & 427 MB & 474 MB \\
\midrule
Acceptance ratios & 2.5\% & 7.5\% & 3\% \\
KenLM memory & 519 MB & 566 MB & 580 MB \\
Baseline total memory & 979 MB & 993 MB & 1054 MB \\
Baseline BLEU & 27.21 & 14.54 & 16.44 \\
\midrule
OxLM memory & 188 MB & 520 MB & 519 MB \\
Compact total memory & 648 MB & 947 MB & 993 MB \\
Compact BLEU & 27.95 & 15.90 & 17.33 \\
\bottomrule
\end{tabular}
\end{center}
\caption{Memory footprints for individual components and overall and end-to-end BLEU scores for the baseline and the compact translation systems across the 3 language pairs.}
\label{table:e2e}
\end{table}

We use the cdec toolkit to train baseline hierarchical phrase-based translation systems using the same parallel and monolingual data. The baseline systems employ traditional back-off n-gram models based on the compact trie implementation available in KenLM \cite{Heafield2011}. In order not to exceed the total 1GB memory limit, the monolingual data used to train the n-gram language models is sampled randomly with the acceptance ratios shown in \autoref{table:e2e}. Phrase tables do not scale for hierarchical phrase based systems and, consequently, the cdec toolkit lacks a phrase table implementation. To overcome this challenge, the baseline systems also rely on the phrase extraction algorithm from \autoref{chapter:extractors} for grammar extraction.

The results of our experiments are shown in \autoref{table:e2e}. The first line shows the amount of memory used by the suffix array grammar extractors. As explained earlier, the same grammar extractors are used by both systems. The next section shows the acceptance ratios used to sample the data for training the n-gram language models, their memory footprint as well as the total footprint of the baseline systems, along with the baseline BLEU scores. The final section shows the memory requirements of the compact systems using the class factored neural language models from \autoref{chapter:nlms} and the corresponding BLEU scores. Our compact systems produce BLEU score improvements of 0.74 for fr$\rightarrow$en, 1.36 for en$\rightarrow$cs and 0.89 for en$\rightarrow$de over the baseline. We note that a larger neural language model could be used for fr$\rightarrow$en (e.g. by projecting the word embeddings and hidden layer into a higher dimensional space) resulting in a higher BLEU score difference, but we decided to share the model configurations among the 3 language pairs to make the results easier to reproduce.

\section{Related Work}

Our work is focused on replacing the memory intensive components of standard machine translation systems in order to make them usable on mobile devices. In recent years, an alternative approach for building translation systems has emerged as the so-called neural translation systems \cite{Kalchbrenner2013,Sutskever2014,Cho2014,Bahdanau2014}. These systems employ a single neural network for modeling the entire translation process unlike standard translation systems which are a collection of purpose specific models (\autoref{chapter:smt}). Neural translation models are an instance of the more generic framework for sequence to sequence mapping \cite{Sutskever2014}, where the models first construct a distributed representation over the source sentence which is then used to generate the target sentence. These models can match state-of-the-art standard translation systems, but require larger and deeper neural architectures than the ones explored in this thesis (e.g. \newcite{Sutskever2014} employ a 4 layer LSTM with 1000 dimensional embeddings) and occupy considerably more memory. A potential solution for this problem is to use several compression tricks introduced for scaling deep architectures for object recognition on mobile devices and embedded systems \cite{Chen2015,Han2015}. \newcite{Chen2015} reduce the amount of redundancy in a neural network by assigning the same value to multiple parameters via hashing and obtain a 8-16x memory reduction rate at no loss in accuracy. \newcite{Han2015} obtain 35-49x reduction rates by pruning the models, applying dynamic weight sharing and using Huffman encoding. These ideas can also be applied to compress the neural language models from \autoref{chapter:nlms}.

Another area of related research is focused on producing compact n-gram language models by pruning weights for n-grams that are unlikely to appear at test time. \newcite{Seymore1996} explore two simple pruning techniques: filtering n-grams below a certain frequency threshold and n-grams whose weight is close to the back-off weight. \newcite{Stolcke2000} proposes a similar approach which aims to minimize the difference in perplexity between the original model and the pruned model. \newcite{Gao2000} construct models for estimating the probability of a n-gram not occurring in the test corpus which they use for pruning the language model. These techniques can be combined with a memory efficient implementation (e.g. the trie based language models available in KenLM \cite{Heafield2011}) to obtain accurate, compact back-off n-gram language models.

\section{Conclusions}

In this thesis, we show how to build a scalable, high quality machine translation system that can operate in memory constrained environments. We present the components in a standard machine translation system (\autoref{chapter:smt}) and  show that phrase tables and n-gram language models are responsible for rendering these systems impractical on devices with limited memory. We propose suffix array grammar extractors and neural language models as compact alternatives for these components.

\autoref{chapter:extractors} discusses suffix array grammar extractors into greater detail. We explain why online extractors are better suited for our goal than storing compact representations of phrase tables on disk. We illustrate how suffix array extractors work for phrase based systems \cite{Lopez2007}. We introduce a novel phrase extraction algorithm for hierarchical phrase based systems which is several times faster than previous work. Our online grammar extractor is open source and has replaced \newcite{Lopez2007}'s implementation as the default extractor in the cdec toolkit. Compared to the previous extractor, one important feature of our implementation is supporting memory efficient parallelism.

\autoref{chapter:nlms} is focused on neural language models. We investigate several strategies for scaling neural language models and their effect on end-to-end translation quality. We experiment with 3 frequently used strategies for reducing the computation in the output layer of the neural network: class factorizations, tree factorizations and ignoring the normalization step altogether. We observe that class factored models are best in terms of translation quality, while unnormalized models are (unsurprisingly) the fastest. The clustering algorithm used to partition the target vocabulary into word classes has considerable impact on quality. In this respect, our experiments show that Brown clustering \cite{Brown1992} clearly outperforms frequency binning. Next, we review noise contrastive training, an effective optimization trick which allows us to skip the normalization step at training time with negligible loss in BLEU score, and show how this technique can be extended to class factored models. We also investigate diagonal context matrices for speeding up the computation of the hidden layer of the neural network. We compare our best performing neural language model with a compact n-gram language model on a wide memory spectrum and show that neural language models are superior when the amount of memory available is limited to 1GB, but if the memory is unbounded, increasing the size of the word embeddings and the hidden layer is not sufficient to obtain parity with n-gram language models. We also investigate two enhancements to neural language models: learning weights for n-grams in the conditioning context jointly with our model and extending the conditioning context to cover windows of relevant source words. These extensions show improvements in translation quality, but also incur an additional memory cost which makes them less suitable for memory constrained environments. We release the OxLM neural language modeling toolkit containing highly optimized implementations of all these models, which can then be integrated as features in the cdec and moses decoders.

In \autoref{chapter:conclusions}, we compare our compact translation system relying on suffix array grammar extractors and class factored neural language models with a baseline hierarchical phase-based translation system trained with the cdec toolkit and show that we obtain considerable gains in BLEU score on 3 language pairs. We review related research and conclude with our findings.

\pagebreak

\addcontentsline{toc}{chapter}{References}

\bibliography{mybib}

\end{document}